\documentclass[10pt,twocolumn,letterpaper]{article}
\usepackage{amsfonts}
\usepackage{authblk}
\usepackage{abstract}
\usepackage[pagenumbers]{cvpr} %

\newif\ifdrafting
\draftingtrue %
\ifdrafting
    \newcommand{\at}[1]{\textcolor{blue}{[AT: #1]}}
    \newcommand{\rr}[1]{\textcolor{red}{[RR: #1]}}
    \newcommand{\kn}[1]{\textcolor{orange}{[KN: #1]}}
    \newcommand{\tw}[1]{\textcolor{green}{[TW: #1]}}
    \newcommand{\sd}[1]{\textcolor{magenta}{[SD: #1]}}

\else
    \newcommand{\at}[1]{}
    \newcommand{\rr}[1]{}
    \newcommand{\kn}[1]{}   
    \newcommand{\tw}[1]{}
    \newcommand{\sd}[1]{}
\fi

\usepackage{multirow}
\usepackage{array}

\usepackage[dvipsnames]{xcolor}

\definecolor{cvprblue}{rgb}{0.21,0.49,0.74}
\usepackage[pagebackref,breaklinks,colorlinks,citecolor=cvprblue]{hyperref}

\def\paperID{534} %
\def\confName{CVPR}
\def\confYear{2024}

\title{What You See is What You GAN: \\Rendering Every Pixel for High-Fidelity Geometry in 3D GANs}

\makeatletter
    \renewcommand\AB@affilsepx{ \hphantom{---} \protect\Affilfont}
\makeatother

\author[2]{Alex Trevithick\thanks{This project was initiated and substantially carried out during an internship at NVIDIA.}}
\author[1]{Matthew Chan}
\author[1]{Towaki Takikawa}
\author[1]{Umar Iqbal}
\author[1]{Shalini De Mello}
\author[2]{Manmohan Chandraker}
\author[2]{Ravi Ramamoorthi}
\author[1]{Koki Nagano}

\affil[1]{NVIDIA}
\affil[2]{University of California, San Diego}

\begin{document}

\twocolumn[{%
\renewcommand\twocolumn[1][]{#1}%
\maketitle
\vspace{-10mm}
\begin{center}
    \centering
    \captionsetup{type=figure}
    \includegraphics[width=\textwidth]{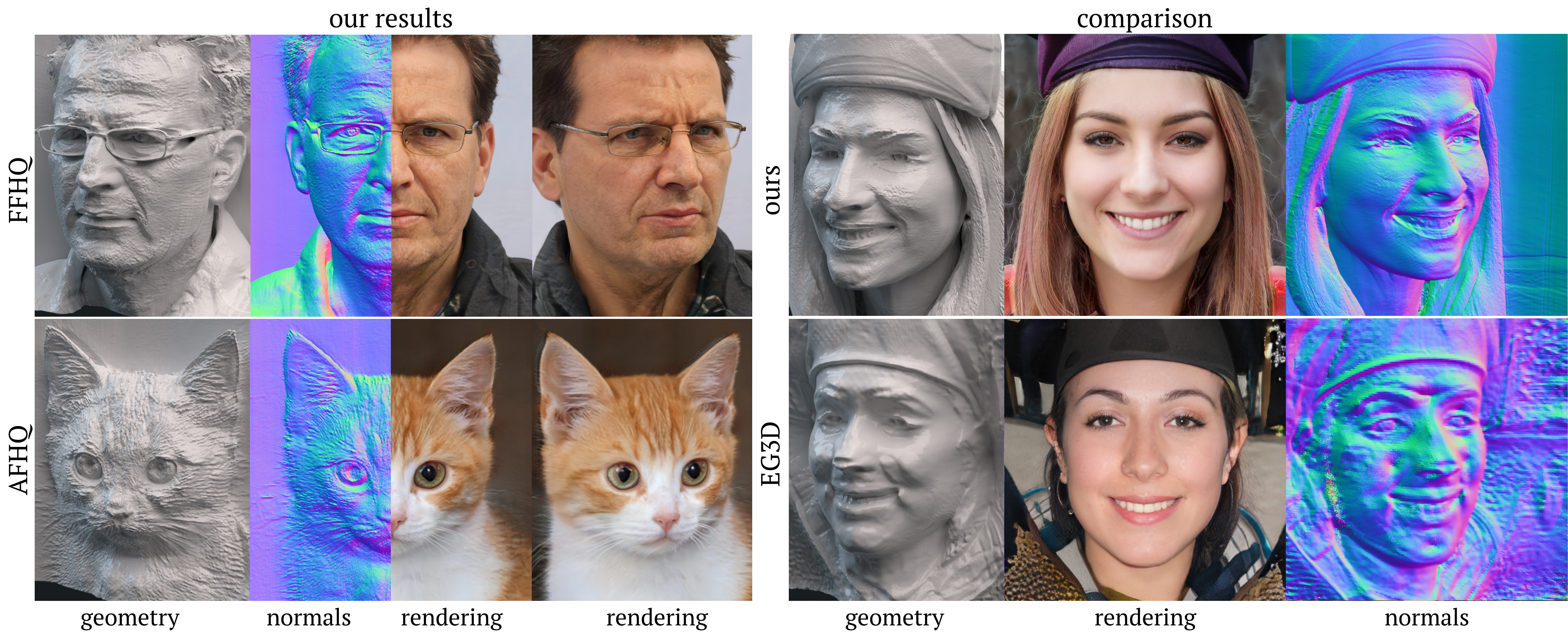}
    \captionof{figure}{Left: Our results. The split view in the middle demonstrates the high degree of agreement between our 2D rendering and corresponding 3D geometry. Our method can learn fine-grained 3D details (e.g., eyeglass frame and cat's fur) that are geometrically well-aligned to 2D images without multiview or 3D scan data. Right: Comparison with EG3D~\cite{eg3d2022}. Our tight SDF prior provides smooth and detailed surfaces on the face and hat while EG3D exhibits geometry artifacts and discrepancies between geometry and rendering. Please see Fig.~\ref{fig:result} and the accompanying video for more examples, and Fig.~\ref{fig:comparison} for comparison to other baselines. }
    \label{fig:teaser}
    \vspace{-0.2cm}
\end{center}
}]

\maketitle

\saythanks

\begin{abstract}
3D-aware Generative Adversarial Networks (GANs) have shown remarkable progress in learning to generate multi-view-consistent images and 3D geometries of scenes from collections of 2D images via neural volume rendering. Yet, the significant memory and computational costs of dense sampling in volume rendering have forced 3D GANs to adopt patch-based training or employ low-resolution rendering with post-processing 2D super resolution, which sacrifices multiview consistency and the quality of resolved geometry. Consequently, 3D GANs have not yet been able to fully resolve the rich 3D geometry present in 2D images.  In this work, we propose techniques to scale neural volume rendering to the much higher resolution of native 2D images, thereby resolving fine-grained 3D geometry with unprecedented detail. Our approach employs learning-based samplers for accelerating neural rendering for 3D GAN training using up to 5 times fewer depth samples. This enables us to explicitly "render every pixel" of the full-resolution image during training and inference without post-processing superresolution in 2D. Together with our strategy to learn high-quality surface geometry, our method synthesizes high-resolution 3D geometry and strictly view-consistent images while maintaining image quality on par with baselines relying on post-processing super resolution. We demonstrate state-of-the-art 3D gemetric quality on FFHQ and AFHQ, setting a new standard for unsupervised learning of 3D shapes in 3D GANs.
\end{abstract}    
\vspace{-.4cm}
\section{Introduction}
\label{sec:intro}
Training 3D generative models from the abundance of 2D images allows the creation of 3D representations of real-world objects for content creation and novel view synthesis~\cite{trevithick2023}. Recently, 3D-aware generative adversarial networks (3D GANs)~\cite{eg3d2022, graf, Niemeyer2020GIRAFFE, xue2022giraffehd, epigraf, gu2021stylenerf, orel2022stylesdf, chan2020pi, Zhou2021CIPS3D, xu2021volumegan,zhang2022mvcgan} have emerged as a powerful way to learn 3D representations from collections of 2D images in an unsupervised fashion. These methods employ differentiable rendering to compare rendered 3D scenes with 2D data using adversarial training~\cite{goodfellow2014generative}. 
Among the various 3D representations, Neural Radiance Fields (NeRF)~\cite{mildenhall2020nerf} have become a popular choice among recent successful 3D GANs. However, the significant computational and memory cost of volume rendering has prevented 3D GANs from scaling to high-resolution output. For instance, generating a single 512x512 image via volume rendering requires evaluating as many as 25 million depth samples, if 96 depth samples are used per ray using importance sampling~\cite{mildenhall2020nerf, epigraf,eg3d2022,mimic}. Given that GAN training typically requires rendering tens of millions of images, the training process could require evaluating \textit{hundreds of trillions} of depth samples. 

During training, all intermediate operations must be stored in GPU memory for every depth sample for the backward pass. Therefore, existing methods resort to working on patches~\cite{graf, mimic, epigraf} or adopting a low resolution neural rendering combined with post-processing 2D super resolution (SR)~\cite{xue2022giraffehd,Niemeyer2020GIRAFFE,eg3d2022,gu2021stylenerf,orel2022stylesdf}. However, patch-based methods have limited receptive fields over scenes, leading to unsatisfactory results, and the hybrid low-resolution rendering and SR scheme inevitably sacrifices the multiview consistency and the accuracy of 3D geometry. While many techniques have been developed to improve the image quality of 3D GANs to match that of 2D GANs, the challenge of resolving the corresponding high-resolution 3D geometry remains unsolved (see Figs.~\ref{fig:teaser} and~\ref{fig:comparison} for our results and comparison to the current state-of-the-art). 

\begin{figure}[t!]
    \centering
    \includegraphics[width=0.7\linewidth]{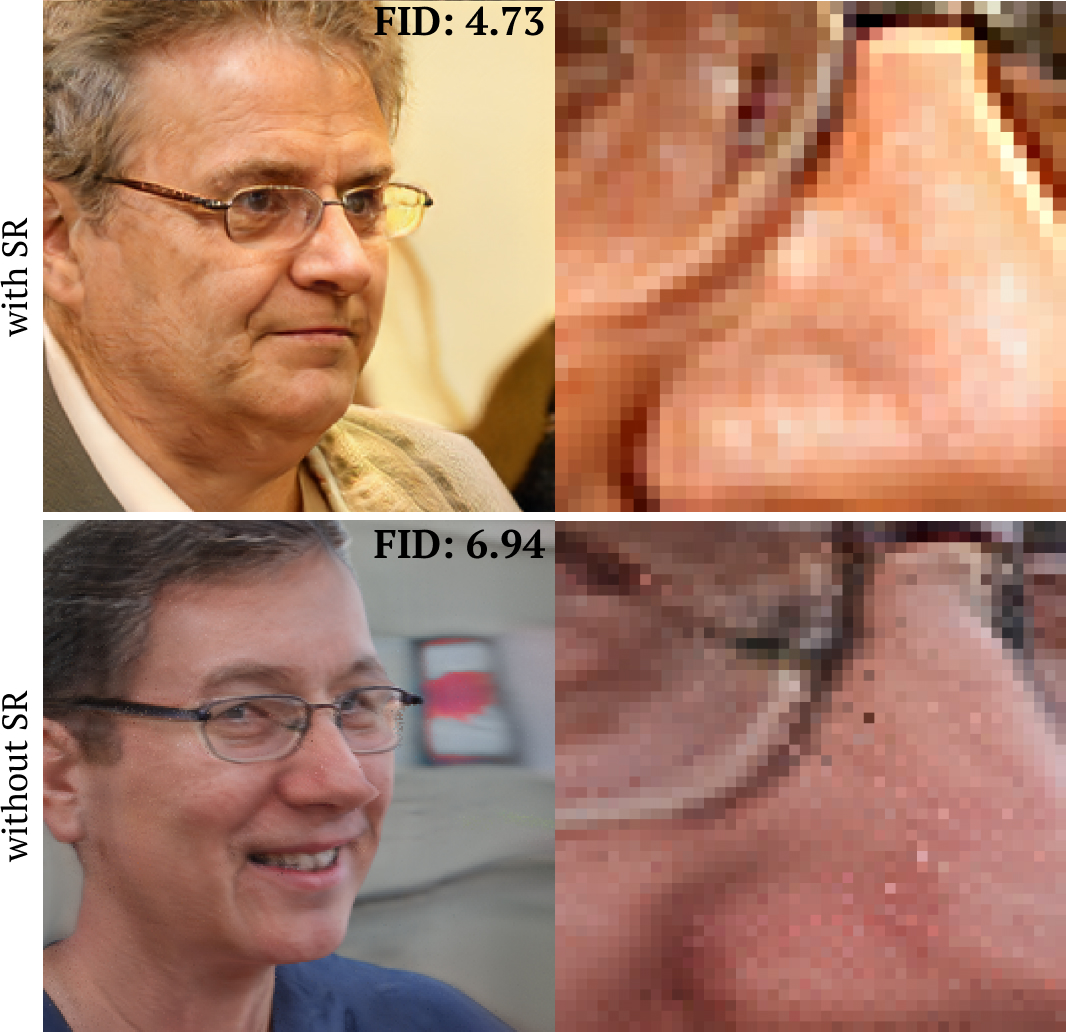}
    \caption{Samples from EG3D 256 model. Right: Volume rendering with 48 coarse samples and 48 fine samples per ray with two-pass importance sampling~\cite{mildenhall2020nerf} results in undersampling, leading to noticeable noisy artifacts. Left: These artifacts are repaired by super resolution (SR). An unsharp mask has been applied to the zoomed views for presentation purposes.}
    \label{fig:eg3d256}
    \vspace{-0.4cm}
\end{figure}

Scaling 3D GANs to operate natively at the 2D pixel resolution requires a novel approach for sampling. Fig.~\ref{fig:eg3d256} compares the state-of-the-art 3D GAN, EG3D model, trained with and without\footnote{Triplane resolution is doubled to compensate for the loss of capacity from removing the SR layers.} SR. EG3D employs 96 dense depth samples in total using two-pass importance sampling~\cite{mildenhall2020nerf} during training, which requires half a terabyte of GPU memory at $256\times256$ resolution, making scaling to higher resolutions infeasible. Furthermore, Fig.~\ref{fig:eg3d256} demonstrates that using 96 dense samples still results in undersampling, as evidenced by the speckle noise patterns visible in the zoomed-in view, leading to considerably worse FID (inset in Fig.~\ref{fig:eg3d256}). 3D GANs relying on post-processing SR layers can repair these undersampling artifacts at the cost of a high-fidelity 3D representation.  

In this work, we address the challenge of scaling neural volume rendering to high resolutions by explicitly rendering every pixel, ensuring that ``\textit{what you see in 2D, is what you get in 3D}'' — generating an unprecedented level of geometric details as well as strictly multiview-consistent images. Our contributions are the following:

\begin{itemize}
    \item We introduce an SDF-based 3D GAN to represent high-frequency geometry with spatially-varying surface tightness that increases throughout training (subsection~\ref{sec:mapping} and~\ref{sec:regularization}), in turn facilitating low sample rendering.
    \item We propose a generalizable learned sampler conditioned on cheap low-resolution information to enable full-resolution rendering \emph{during training} for the first time (subsection~\ref{sec:probe} and ~\ref{sec:supervision}). 
    \item We show a robust sampling strategy for the learned sampler (subsection~\ref{sec:sampling}) that produces stable neural rendering using significantly fewer depth samples (see Fig.~\ref{fig:ablation}). Our sampler can operate with just 20 samples per ray compared to existing 3D GANs which must use at least 96 samples per ray (see Table~\ref{table:20spp}).
    \item Together, our contributions result in the state-of-the-art geometry for 3D GANs while rendering with quality on par with SR baselines (see Fig.~\ref{fig:teaser}). For more results, see Fig.~\ref{fig:result} and for comparison to other baselines, see Fig.~\ref{fig:comparison}.
\end{itemize}

\section{Related Work}

We begin by reviewing the prior-arts of 3D generative models and their current shortcomings. We then cover foundational techniques for 3D geometry representation and neural rendering from which we take inspiration. We then discuss existing methods for accelerating neural volume rendering, which usually operate \textit{per-scene}.

\subsection{3D Generative Models}
\label{subsec:related-3dgan}
Just like 2D GANs, 3D-aware GANs train from a collection of 2D images, but employ a 3D representation and differentiable rendering to learn 3D scenes without requiring multi-view images or ground truth 3D scans. 
Some of the most successful works use a neural field~\cite{Yiheng2022neuralfield} in combination with a feature grid~\cite{eg3d2022} as their 3D representation, and use neural volume rendering~\cite{mildenhall2020nerf} as the differentiable renderer. 

However, due to the significant memory and computational cost of neural volume rendering, many previous works perform rendering at low-resolution and rely on a 2D post-processing CNN~\cite{xue2022giraffehd,Niemeyer2020GIRAFFE,eg3d2022,gu2021stylenerf,orel2022stylesdf}, which \textit{hallucinates} the high-frequency details in a view-inconsistent manner while sacrificing 3D consistency and the quality of the resolved 3D geometry.

To ensure strict 3D consistency, other previous works seek to render at high-resolutions and propose techniques to address the prohibitive computational costs. One line of work leverages the sparse nature of 3D scenes to speed up rendering, in particular, structures such as 2D manifolds \cite{deng2022gram,xiang2022gramhd}, multiplane images~\cite{zhao-gmpi2022} and sparse voxels~\cite{voxgraf}. 
Although they are more efficient, sparse representations provide only coarse~\cite{voxgraf} or category-specific~\cite{deng2022gram,xiang2022gramhd} acceleration structures, which poses constraints on the diversity and viewing angles of the generated scenes. 
Our sampling-based method, on the other hand, generalizes to every new scene and adaptively accelerates rendering on a per ray basis.
Another line of work enables high-resolution rendering with patch-based training~\cite{epigraf,mimic}. In particular, Mimic3D~\cite{mimic} achieves significantly improved 2D image quality, 
but the patch-based training limits the receptive fields, and the generated geometry does not faithfully represent the 2D data due to the patch-wise perceptual loss. Our method renders the entire image at once and the resulting geometry is aligned with the rendering (see Figs.~\ref{fig:teaser} and~\ref{fig:comparison}).

Recently, a new family of generative approaches using diffusion models has been proposed to tackle conditional tasks including novel view synthesis~\cite{chan2023genvs,tewari2023diffusion} and text-based 3D generation\cite{Wang_2023_CVPR_rodin}. Most of these 3D-aware diffusion models combine a 3D inductive bias modeled via neural field representations and a 2D image denoising objective to learn 3D scene generation. While these models enable unconditional 3D generation, they require multiview images~\cite{szymanowicz23viewset_diffusion, karnewar2023holodiffusion,Wang_2023_CVPR_rodin} or 3D data, such as a point cloud~\cite{nichol2022pointe}. Score Distillation Sampling~\cite{poole2022dreamfusion} may be used for distillation from a pre-trained 2D diffusion model when only monocular 2D data is available, but diffusion models incur significant computational costs due to their iterative nature and most of the existing methods require optimization per scene~\cite{sjc, vsd, fantasia, magic3d, nerfdiff}. 

\subsection{Learning High-Fidelity Geometry}
\label{subsec:related-geometry}

Prior works on 3D GANs have typically represented the geometry as a \textit{radiance field}~\cite{mildenhall2020nerf}, which lacks a concrete definition for where the surface geometry resides in the field, resulting in bumpy surfaces. A number of works~\cite{yariv2020multiview,oechsle2021unisurf,wang2021neus} have proposed alternate representations based on implicit surfaces (such as signed distance functions, or SDFs) that can be used with neural volume rendering. In these works, the implicit surface is typically \textit{softened} by a parameter for volume rendering. 

Other works~\cite{neus2, li2023neuralangelo} improve on these implicit surface representations by leveraging feature grids for higher computational efficiency and resolution. Adaptive Shells~\cite{zian2023adaptiveshells} further improve on quality by making the \textit{softness} parameter spatially-varying, as many objects have hard boundaries only in certain parts. We use an implicit surface representation based on VolSDF~\cite{yariv2021volsdf}, and leverage a spatially-varying parameter similar to Adaptive Shells~\cite{zian2023adaptiveshells} to control the softness of the surface, as humans and animals benefit from both hard surfaces (e.g. skin, eyes) and \textit{soft}, volumetric representations (e.g. hair). Although other works such as StyleSDF~\cite{orel2022stylesdf} have similarly leveraged implicit surfaces in a 3D GAN framework, the lack of spatial-variance and high-resolution rendering led to over-smoothed geometry not faithful to the rendered images.

\subsection{Accelerating Neural Volume Rendering}
\label{subsec:related-acceleration}

As mentioned in Section~\ref{subsec:related-3dgan}, accelerating 3D GANs typically relies on acceleration structures such as  octrees~\cite{yu2021plenoctrees,takikawa2021nglod,li2023nerfacc}, which in generative settings~\cite{voxgraf,deng2022gram,xiang2022gramhd} are limited to be coarse or category-specific due to the difficulty of making them generalize \textit{per-scene}. Instead, we look to a class of methods that do not rely on an acceleration structure. Some of these works learn a per-scene sampling prior on a per-ray basis using a binary classifier~\cite{neff2021donerf}, a density estimator~\cite{kurz-adanerf2022, terminerf}, a coarse proxy from a 3D cost volume~\cite{lin2022efficient}, or an interval length estimator~\cite{lindell2020autoint} on discrete depth regions along the ray. Other works use importance sampling~\cite{mildenhall2020nerf, gupta2023mcnerf} to sample additional points. We take inspiration from works on density estimators~\cite{kurz-adanerf2022} and propose to learn a \textit{scene-conditional} proposal network that generalizes across scenes instead of being category-specific or optimized per-scene.

There are also other methods to accelerate rendering by utilizing more efficient representations, such as gaussian splats~\cite{kerbl3Dgaussians} and light field networks~\cite{sitzmann2021light}. More efficient feature grids~\cite{muller2022instant, takikawa2023compact} based on hashing can also be used to accelerate rendering. However, mapping to these representations in a GAN framework is not straightforward. In contrast, our sampling strategy can be used for any NeRF representation.

    \vspace{-0.3cm}

\label{sec:related}

\begin{figure*}
    \centering
    \includegraphics[width=\linewidth]{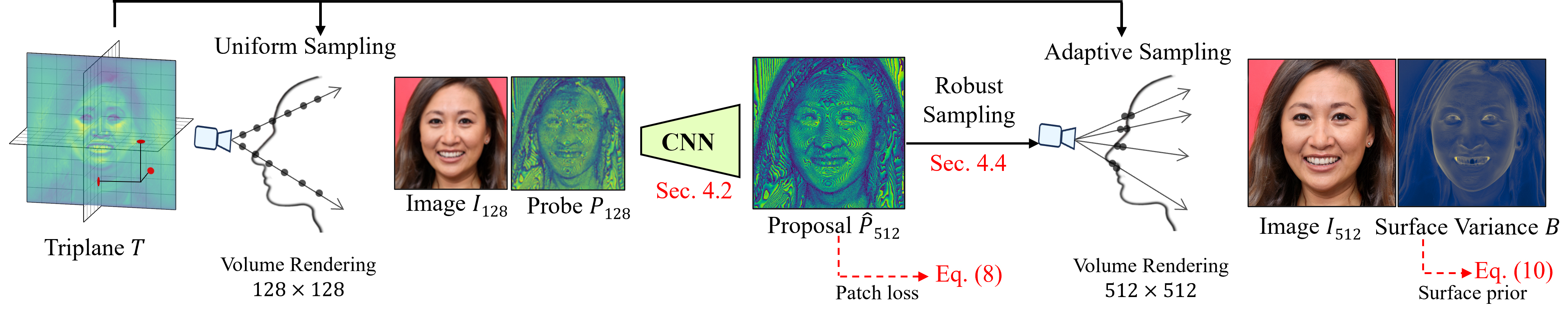}
    \caption{Here we show our proposed pipeline and its intermediate outputs. Beginning from the triplane $T$, we trace uniform samples to probe the scene, yielding low-resolution $I_{128}$ and weights $P_{128}$. These are fed to a CNN which produces high-resolution proposal weights $\hat P_{512}$ (weights are visualized as uniform level sets). We perform robust sampling and volume render to get the final image $I_{512}$ and the surface variance $B$. }
    \label{fig:pipeline}
\vspace{-.4cm}
\end{figure*}

\section{Background}
\label{sec:background}
We begin with background on the methodology of the state-of-the-art 3D-aware GANs as our method relies on a similar backbone for mapping to 3D representations. 3D GANs typically utilize a StyleGAN-like~\cite{Karras2020stylegan2} architecture to map from a simple Gaussian prior to the conditioning of a NeRF, whether that be an MLP~\cite{gu2021stylenerf, orel2022stylesdf}, MPI~\cite{zhao-gmpi2022}, 3D feature grid~\cite{voxgraf}, manifolds~\cite{deng2022gram,xiang2022gramhd} or triplane~\cite{eg3d2022, mimic, epigraf}. We inherit the latter triplane conditioning for its high expressivity and efficiency, in which three axis-aligned 2D feature grids ($f_{xy}, f_{xz}, f_{yz})$, provide NeRF conditioning by orthogonal projection and interpolation. As in the previous methods, the mapping and synthesis networks from StyleGAN2 can easily be adapted to create the 2D triplane representation from noise $z\in\mathbb{R}^{512}$. Specifically, $\mathbf{w}=\text{Mapping}(z)$ conditions the creation of the triplane $T(\mathbf{w})\in\mathbb{R}^{3\times 32\times 512\times 512}$ from a Synthesis network, corresponding to three axis-aligned 2D feature grids of spatial resolution $512\times512$ and feature size $32$.

To create high-fidelity geometry, our method builds upon VolSDF~\cite{yariv2021volsdf}: Instead of directly outputting the opacity $\sigma$ of a point $x\in\mathbb{R}^3$, an SDF value $s$ is output and transformed to $\sigma$ using a Laplacian CDF: 

\begin{equation}
\label{equation:laplace}
\sigma = \frac{1}{\beta} 
\begin{cases} 
\frac{1}{2} \exp\left(\frac{s}{\beta}\right) & \text{if } s \leq 0 \\
1 - \frac{1}{2} \exp\left(-\frac{s}{\beta}\right) & \text{if } s > 0 
\end{cases}
\end{equation}
where $\beta$ is the variance of the Laplacian distribution governing the ``tightness" of the representation. One distinct benefit of an SDF-based representation is the ease of extracting the surface. StyleSDF~\cite{orel2022stylesdf} also utilizes this intermediate geometry representation without a triplane, enforcing the usual Eikonal constraint.

Using the triplane conditioning and Eq.~\ref{equation:laplace}, we can assign each point in the volume with its opacity $\sigma$ and radiance $c$ using a lightweight MLP. For a given camera ray $r(t) = o + td$, we approximate the volumetric rendering integral $C(r)$~\cite{Max:1995} by sampling ray distances $t_i$ with their corresponding $\sigma_i$ and $c_i$ before computing
\begin{equation}
\label{equation:vr}
\begin{aligned}
\hat{C}(r) = \sum_{i=1}^{N} w_i c_i \text{ ~where~ } w_i = T_i (1 - \exp(-\sigma_i \delta_i)), \\
T_i = \exp\left(-\sum_{j=1}^{i-1} \sigma_j \delta_j\right) \text{ and } \delta_i = t_{i+1} - t_{i}.
\end{aligned}
\end{equation}
Here, $T_i$ denotes the accumulated transmittance and $\delta_i$ is the distance between adjacent samples along the ray. 

It is possible to develop a more efficient estimator for this sum with fewer samples by using importance sampling techniques in computer graphics. Typically, one computes a piecewise constant probability distribution $p_j = \frac{\hat{w}_j}{\sum_j \hat{w}_j}$, where $j$ refers to the $j^{\mathrm{th}}$ bin or region, and $\hat{w_j}$ is an estimate of $w_j$ for that region, for example obtained by explicitly tracing coarse samples.  For a given $t$, we first find the region $j(t)$ and then set $p(t) = p_j$.  From this, one can compute a (piecewise linear) cumulative distribution function or CDF $\Phi(t)$ which has a range from $0$ to $1$.  We can then perform inverse CDF sampling to define the sample points, 
\begin{equation}
\label{equation:inverse_cdf}
t_i = \Phi^{-1}(u_i),
\end{equation}
where $u_i$ is a random number from $0$ to $1$ (sorted to be an increasing sequence). 
\vspace{-.45cm}
\paragraph{Discussion}
We improve on previous works such as NeRF~\cite{mildenhall2020nerf} and EG3D~\cite{eg3d2022} by stratifying\footnote{dividing the unit interval into bins and taking a sample from each bin.} the random numbers $u_i$ during training; this leads to significantly lower rendering variance, especially at low sample counts $N$~\cite{mitchell1996consequences}.  We also develop a neural method to predict a good distribution $p_j$ (and hence $\Phi$) for importance sampling at high spatial resolution, without needing to exhaustively step through the ray.

\section{Method}
\label{sec:method}
In this section, we describe our method beginning with our SDF-based NeRF parametrization (subsection~\ref{sec:mapping}). We then overview how we render at high-resolution in three stages: first, a low-resolution probe into the 3D scene (subsection~\ref{sec:probe}); second a high-resolution CNN proposal network (subsection~\ref{sec:supervision}); and third a robust sampling method for the resultant proposals (subsection~\ref{sec:sampling}). Next we describe regularizations (subsection~\ref{sec:regularization}) for stable training, and finally our entire training pipeline (subsection~\ref{sec:overview}).

\subsection{Mapping to a 3D Representation}
\label{sec:mapping}
Beginning from a noise vector $z$, we synthesize the initial triplane $T'$ with StyleGAN~\cite{karras2019style} layers as detailed in Sec. \ref{sec:background}. In contrast to previous methods, we then generate more expressive triplane features $T$ with an extra synthesis block for each orthogonal plane: $f_{ij} = \text{SynthesisBlock}(f'_{ij})$ where $ij\in\{xy,xz,yz\}$ and $f'$ are the features of $T'$. This design choice allows disentanglement between the intermediate triplane features as plane-specific kernels can attend to the features in each separate plane.

Given the triplane $T$ (the left side of Fig.~\ref{fig:pipeline}) and a point $x\in\mathbb{R}^3$, we utilize an MLP to map to the SDF value $s$, variance $\beta$ and geometry features $f_{geo}$:
\begin{equation}
    (s,\beta,f_{geo}) = \text{MLP}_\text{SDF} (\text{PosEnc}(x), T_x)
\end{equation}
where $\text{PosEnc}$ is the positional encoding from NeRF~\cite{mildenhall2020nerf} and $T_x$ are the features corresponding to $x$ gathered from $T$ by projecting $x$ to each of the axis-aligned planes and taking the Hadamard product of the three resultant vectors~\cite{kplanes}. We initialize the ReLU $\text{MLP}_\text{SDF}$ as in SAL~\cite{sal} to ensure an approximately spherical SDF in the early part of training. Additionally note that $\beta$ varies spatially in the volume unlike ~\cite{yariv2021volsdf,orel2022stylesdf}, allowing us to regularize its values later on.

Using Eq.~\ref{equation:laplace}, we transform $s$ and $\beta$ into opacity $\sigma$. We can now predict the radiance with a separate $\text{MLP}_\text{c}$ conditioned on the geometry features and viewing direction $v$ as 
\begin{equation}
\label{equation:MLP_c}
    c = \text{MLP}_\text{c}(\text{PosEnc}(v), f_{geo}).
\end{equation}

Note that in contrast to most 3D GANs, we condition radiance on the viewing direction, allowing a more expressive generator. Thus, given a triplane, we can render any pixel by computing $\sigma$ and $c$ for points along the ray to approximate the volumetric rendering integral as described in Sec.~\ref{sec:background}.

\subsection{High-Resolution Proposal Network} \label{sec:probe}
We now have our mapping from a latent code $z$ to a 3D NeRF representation. However, volumetric rendering of NeRFs at higher resolutions requires extremely large numbers of samples, and thus both memory and time. Instead of naive dense sampling at a high resolution, we propose to leverage low-resolution renderings to cheaply probe the 3D representation (visualized on the left of Fig.~\ref{fig:pipeline}) for the creation of proposal distributions at high-resolution. Given a target camera and triplane $T$, we first trace $192$ coarse samples at low-resolution ($128\times128$) to compute a low-resolution RGB image $I_{128}\in\mathbb{R}^{3\times128\times128}$ and a tensor of weights $P_{128}\in\mathbb{R}^{192\times128\times128}$ (visualized after low-resolution rendering in Fig.~\ref{fig:pipeline}). Each $192$-dimensional vector corresponds to a piecewise constant PDF with CDF $\Phi$ as seen in Eq.~\ref{equation:inverse_cdf}.

Conditioned on the low-resolution probe, we predict a tensor of proposal volume rendering weights at the high-resolution ($512\times 512$):
\begin{equation}
    \hat P_{512} = \text{Softmax}(\text{CNN}(P_{128}, I_{128})) \in\mathbb{R}^{192 \times 512 \times 512},
\end{equation}
where CNN is a lightweight network that up-samples the low-resolution    weights, Softmax produces discrete distributions along each ray, and the $\hat{\phantom{a}}$ denotes that this is an estimated quantity. This corresponds to the Proposal in Fig.~\ref{fig:pipeline} and the yellow distribution in Fig.~\ref{fig:pdf}. Note that allocating $192$ samples at $128\times128$ is equivalent to allocating just $12$ at $512\times 512$.

\subsection{Supervising the Proposal Network}\label{sec:supervision}
Having described the input and output of our high-resolution proposal network, we now show its supervision. From the target camera, we can also trace $192$ coarse samples at high resolution for a small $64\times 64$ patch, giving us a ground truth tensor of volume rendering weights $P_{\text{patch}}\in\mathbb{R}^{192 \times 64 \times 64}$. We then prepare this tensor for supervision by computing:
\begin{equation}
    \bar P_{\text{patch}} = \text{Normalize}( \text{Suppress} ( \text{Blur} (P_{\text{patch}})))
\end{equation}
where Blur applies a 1D Gaussian kernel to the input distributions, Suppress$(x)=x$ if $x\geq5e-3$ and $0$ otherwise, and Normalize is L1 normalization to create a valid distribution. This corresponds to the patch loss in Fig.~\ref{fig:pipeline} and the purple distribution in Fig.~\ref{fig:pdf}.  These operations create less noisy distributions to facilitate accurate learning of the high-frequency integrand which may be undersampled in the coarse pass. 

We can then compare the predicted and cleaned ground truth distributions with a cross-entropy loss:
\begin{equation}
    L_\text{sampler} = \text{CrossEntropy}(\bar P_{\text{patch}}, \hat P_{\text{patch}})
\end{equation}
where $\hat P_{\text{patch}}$ is the corresponding patch of weights in $\hat P_{512}$ and CrossEntropy denotes the average cross-entropy between all pairs of pixelwise discrete distributions; for each pair $(\bar p,\hat p)$, we compute $\sum -\bar p_j\log \hat p_j$. Since we only need to compute this supervision for a small patch, the overhead of sampler training is not significant.

\subsection{Sampling from the Proposal Network}\label{sec:sampling}
\begin{figure}[t!]
    \centering
    \includegraphics[width=\linewidth]{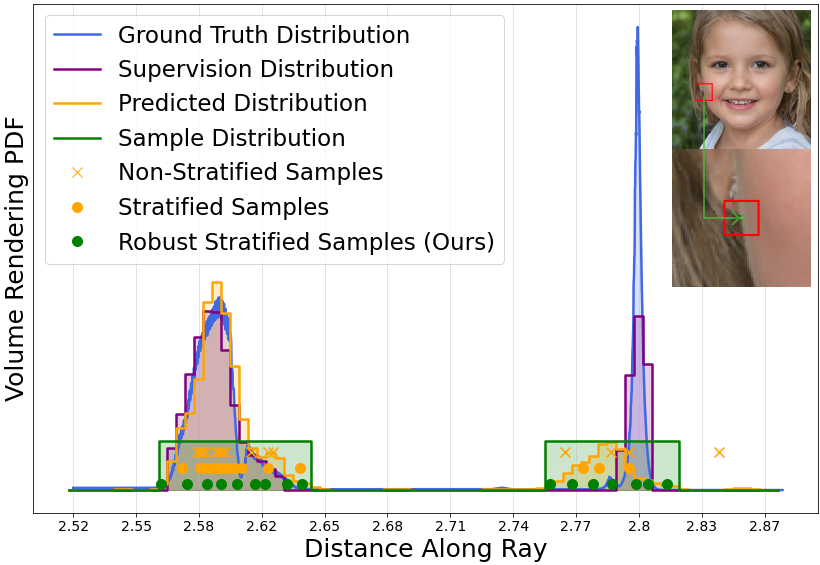}
    \caption{We visualize the volume rendering PDFs for the green pixel in the images on the right along with sampling methods. The ground truth distribution in blue is bimodal due to the discontinuous depth. Without stratification, the samples from the predicted yellow PDF completely miss the second mode. Stratification reduces the variance, yet also misses the second mode. Our robust stratified samples hit both modes despite the inaccurate predictions. The supervision PDF is visualized in purple as well. }
    \label{fig:pdf}
        \vspace{-0.3cm}
\end{figure}

Having shown how to train and predict high-resolution distributions, we now overview how to \emph{sample} the resultant proposals. As seen in Fig.~\ref{fig:pdf}, the proposals are often slightly off; this is due to the high frequency nature of the underlying integrand in blue. 

In order to utilize the information from the sampler, we propose to filter predicted PDFs for better estimation. Specifically, for each discrete predicted PDF $\hat p$, we compute the smallest set of bins whose probability exceeds a threshold $\tau=0.98$: We find the smallest subset $I\subseteq \{1,2,...,192\}$ such that 
\begin{equation}
    \sum_{i\in I} \hat p_i \geq \tau.
\end{equation}
This operation resembles nucleus sampling in NLP~\cite{topp}. We define our sampling PDF $q$ with probability $q_i=\textstyle{\frac{1}{|I|}}$ if $i\in I$ and $0$ otherwise (the green distribution in Fig.~\ref{fig:pdf}). 

For each PDF $q$, we compute its CDF $\Phi$ and perform stratified inverse transform sampling to create the samples (illustrated as adaptive sampling near the surface in Fig.~\ref{fig:pipeline}). In practice, on top of the $12$ samples from the coarse probe (for the high-resolution image), we take an additional $18$ samples per pixel adaptively based on the variance of the predicted distributions. The details are given in the supplement.

\begin{figure*}
    \centering
    \includegraphics[width=.97\textwidth]{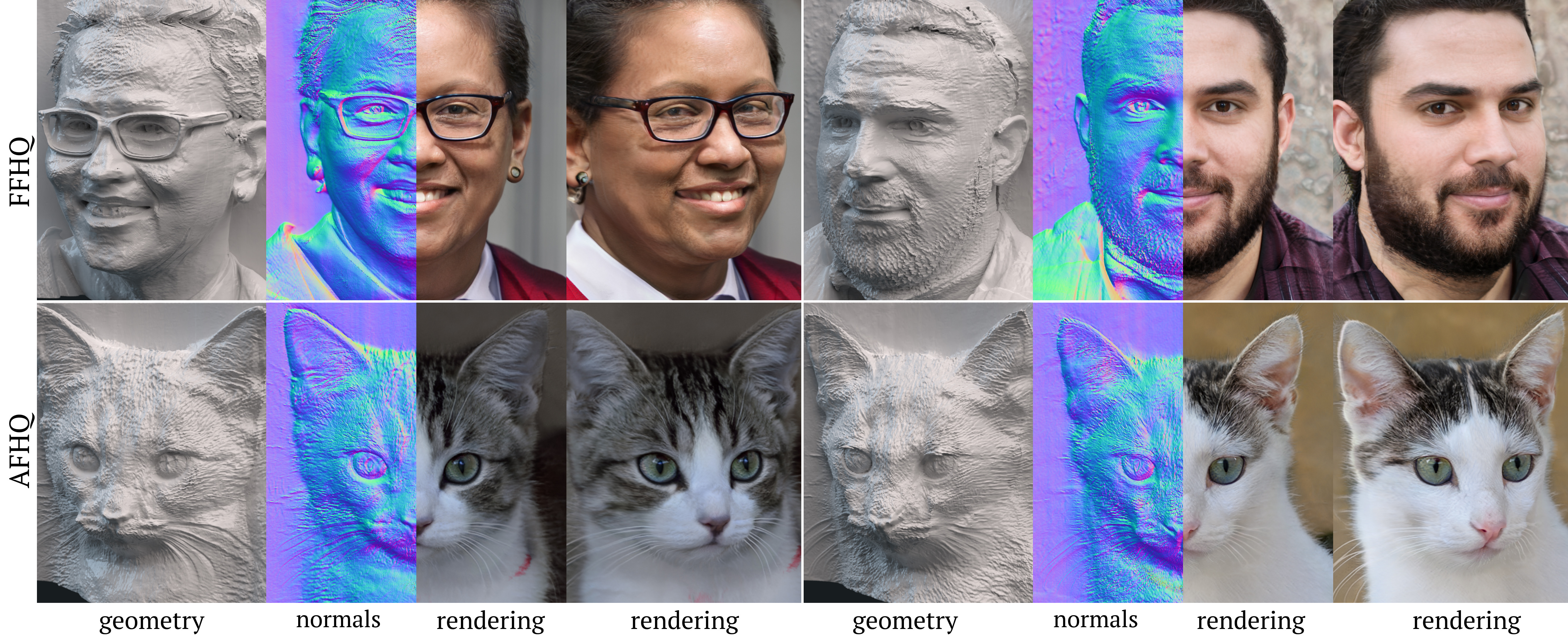}
    \caption{Curated samples on FFHQ and AFHQ. Our method can resolve high-fidelity geometry (e.g., eyeglasses) and fine-grained details (e.g., stubble hair and cat's fur) as seen in the geometry and normal map.}
    \vspace{-0.3cm}
    \label{fig:result}
\end{figure*}

\subsection{Regularization for High-Resolution Training}
\label{sec:regularization}
\begin{figure}[t!]
    \centering
    \includegraphics[width=\linewidth]{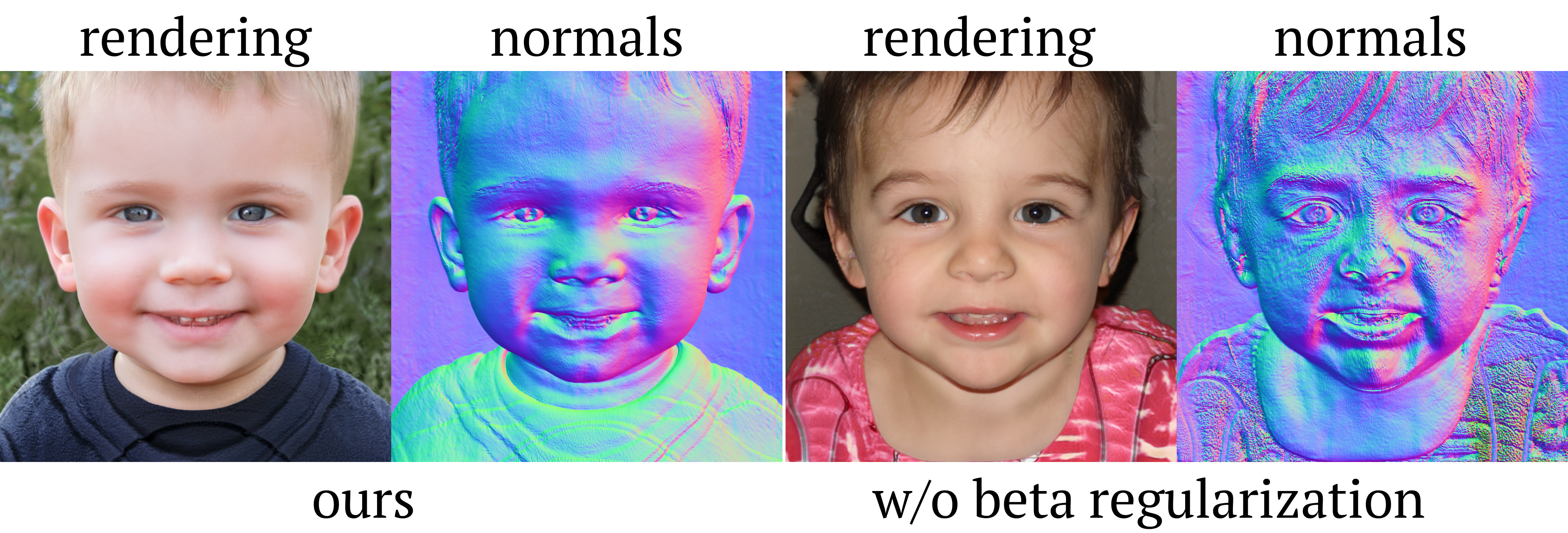}
    \caption{Ablation study on the effect of beta regularization.}
    \label{fig:beta}
    \vspace{-0.5cm}
\end{figure}

In order to render accurately under low sample budget per ray, we desire the surface to be tight, i.e., the set of points along the ray, which contribute to the accumulated radiance to be small. To accomplish this, we introduce a regularization for the spatially-varying $\beta$ values. Replacing the $c_i$ with the intermediate $\beta_i$ in Eq.~\ref{equation:vr}, we can volume render an image of scalars, $B\in\mathbb{R}^{512\times512}$ (seen on the right of Fig.~\ref{fig:pipeline}). During training, we regularize $B$ to be small so that the surface tightens:
\begin{equation}
\label{equation:beta}
    L_\text{surface} = \sum_{hw}(B_{hw} - B_\text{target})^2,
\end{equation}
where $B_\text{target}$ is a scalar quantity annealed towards $\epsilon>0$ during optimization. Note that this results in significantly more pronounced and smooth geometry (see Fig.~\ref{fig:beta}).

During training with the proposal network, the rendering networks only receive gradients for points very near the surface (which have high density). We find that this can lead to undesirable density growing from the background into the foreground. In order to combat this, we leverage the low-resolution information from $I_{128}$, which is computed for uniform samples along the ray, and thus not concentrated at the surface. Considering the SDF values intermediately computed for rendering, $S=S_{128}\in \mathbb{R}^{192\times 128\times128}$, we enforce the SDF decision boundary by minimizing the SDF likelihood under a Laplacian distribution similar to~\cite{sitzmann2020siren, orel2022stylesdf}:
\begin{equation}
    L_{\text{dec}}  = \sum_{zhw} \exp{\left( - 2|S_{zhw}| \right)}.
\end{equation}

 \subsection{The Training Pipeline}
\label{sec:overview}
 In order to begin making use of the learned sampler for high-resolution rendering, we need a good NeRF representation from which to train it. Concurrently, we also need a good sampler in order to allow NeRF to render at $512\times 512$, the input resolution to the discriminator $D$.

To solve this issue, in the early stages of training, we first learn a low-resolution (e.g., $64\times64$) 3D GAN through the standard NeRF sampling techniques. We bilinearly upsample our low-resolution renderings to $512\times512$ and blur the real images to the same level. After converging at the lower resolution, we introduce the sampler training (subsection~\ref{sec:supervision}). Concretely, we not only render low-resolution images with standard sampling, but also render sampler inputs $P_{128}$ and supervision patches $P_\text{patch}$. This results in a good initialization for the sampler. 

Having learned an initial low-resolution 3D GAN and high-resolution sampler, we transition to rendering with the sampler predictions. The high-resolution proposals $\hat P_{512}$ are downsampled to the current rendering resolution, which is progressively increased to the full $512\times 512$ resolution during training. After introducing all losses, we optimize the parameters of the generator $G$ to minimize the following:
\begin{equation}
     L = L_\text{adv} + \lambda_\text{sampler}L_\text{sampler} +\lambda_\text{surface}L_\text{surface} + \lambda_\text{dec}L_\text{dec} 
\end{equation}
where $L_\text{adv}$ is the standard GAN loss~\cite{goodfellow2014generative} $\text{Softplus}(-D(G(z))$. Note that we do not enforce the Eikonal constraint as we did not see a benefit. The discriminator $D$ is trained with R1 gradient regularization~\cite{r1} whose weight is adapted as the resolution changes. Generator and discriminator pose conditioning follow EG3D~\cite{eg3d2022}. The details of all hyperparameters and schedules are presented in the supplementary material.

\begin{figure*}
    \centering
    \includegraphics[width=.97\textwidth]{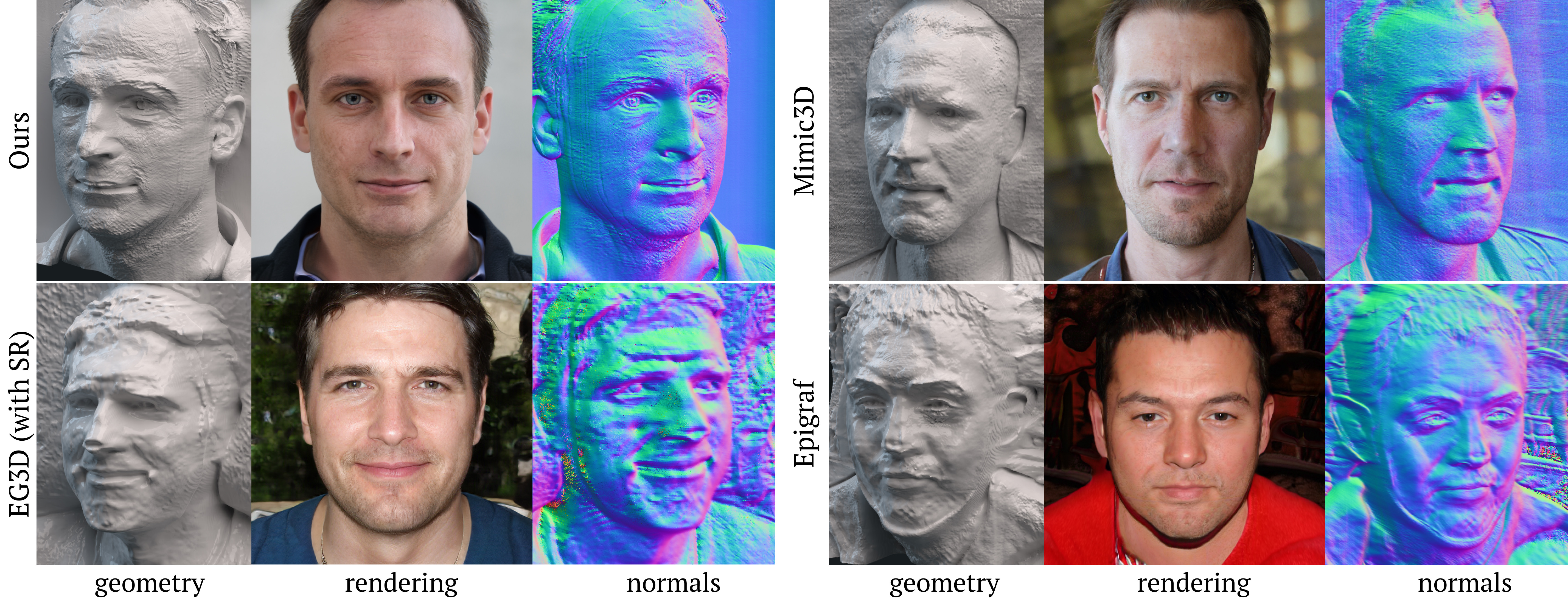}
    \caption{Qualitative comparisons on FFHQ with EG3D~\cite{eg3d2022}, Mimic3D~\cite{mimic} and Epigraf~\cite{epigraf}.  EG3D performs neural rendering at resolution $128\times 128$ and relies on $4\times$ super resolution to generate images. On the right, Mimic3D and Epigraf directly generate the image via neural rendering. While all other baselines use up to 192 dense depth samples per ray, our method can operate at 30 samples per ray.}
    \label{fig:comparison}
    \vspace{-0.3cm}
\end{figure*}

\section{Results}
\paragraph{Datasets.}
We benchmark on two standard datasets for 3D GANs: FFHQ~\cite{karras2019style} and AFHQv2 Cats~\cite{choi2020starganv2, Karras2021} both at resolution $512\times 512$. We use the camera parameters extracted by EG3D~\cite{eg3d2022} for conditioning and rendering. Our AFHQ model is finetuned from our FFHQ model using adaptive data augmentation~\cite{Karras2020ada}. For more results, please see the accompanying video. 
\subsection{Comparisons}
\paragraph{Baselines.} We compare our methods against state-of-the-art 3D GAN methods including those that use low-resolution neural rendering and 2D post-processing CNN super resolution: EG3D~\cite{eg3d2022}, MVCGAN~\cite{zhang2022mvcgan}, and StyleSDF~\cite{orel2022stylesdf}; and methods that operate entirely based on neural rendering: Mimic3D~\cite{mimic}, Epigraf~\cite{epigraf}, GMPI~\cite{zhao-gmpi2022}, and GramHD~\cite{xiang2022gramhd}. 
\vspace{-.3cm}
\paragraph{Qualitative results}
Fig.~\ref{fig:result} shows the curated samples generated by our method with FFHQ and AFHQ, demonstrating photorealistic rendering as well as high-resolution details that align with the 2D images. Fig~\ref{fig:comparison} provides qualitative comparisons to baselines. EG3D shows significant artifacts and cannot resolve high-resolution geometry since it performs neural rendering at only $128\times128$. Mimic3D and Epigraf render all pixels with neural rendering, but the patch-based nature of these methods harm the overall 3D geometry (e.g., distorted face and missing ear). Our method provides both high-fidelity 3D shape (e.g., well-defined ears and isolated clothing collar) and high-resolution details.
\vspace{-.3cm}
\paragraph{Quantitative results.}
Tabs.~\ref{table:FFHQ} and ~\ref{table:AFHQ} provide quantitative comparisons against baselines. We measure the image quality with Fr\'echet Inception Distance ({FID})~\cite{DBLP:journals/corr/HeuselRUNKH17}. We assess the quality of the learned geometry with a face-specific Normal FID (FID-N)~\cite{ag3d}. We render 10.79k normal maps from the NPHM~\cite{nphm} dataset by solving for the approximate alignment between the average FFHQ 2D landmarks and the provided 3D landmarks. Examples are given in the supplement. For each mesh, we render two views with approximately 20 degrees of pitch variation relative to the front of the face. These views are processed to remove the background with facer~\cite{facer}. For each baseline, we render 10k normal maps and remove their background using the predicted mask from facer on the rendered image. We compute the FID between the two sets of images. Finally, we also evaluate the flatness of the learned representations with non-flatness score (NFS)~\cite{imagenet3d}.

Our results show the state-of-the-art image quality among the methods that operate only with neural rendering, while achieving FID comparable to the state-of-the-art SR-based method, EG3D. Our geometry quality outperforms all existing methods as indicated by our state-of-the-art FID-N. Additionally, our high NFS scores show the 3D aspect of our geometry. However, since NFS simply measures the variations of depth as a measure of non-flatness, it does not quantify the quality of geometry above a certain threshold.  

We also compare other methods' ability to render with low sample count in Tab.~\ref{table:20spp}. With just 20 samples, the rendering quality of our method drops by only $.3$ in FID, compared to the precipitous drop for other methods. This validates our strategy to jointly learn a sampler and ensure tight SDF surface for operation under a limited sample budget.
\vspace{-.1cm}

\subsection{Ablation Study}
\begin{figure}[t!]
    \centering
    \includegraphics[width=\linewidth]{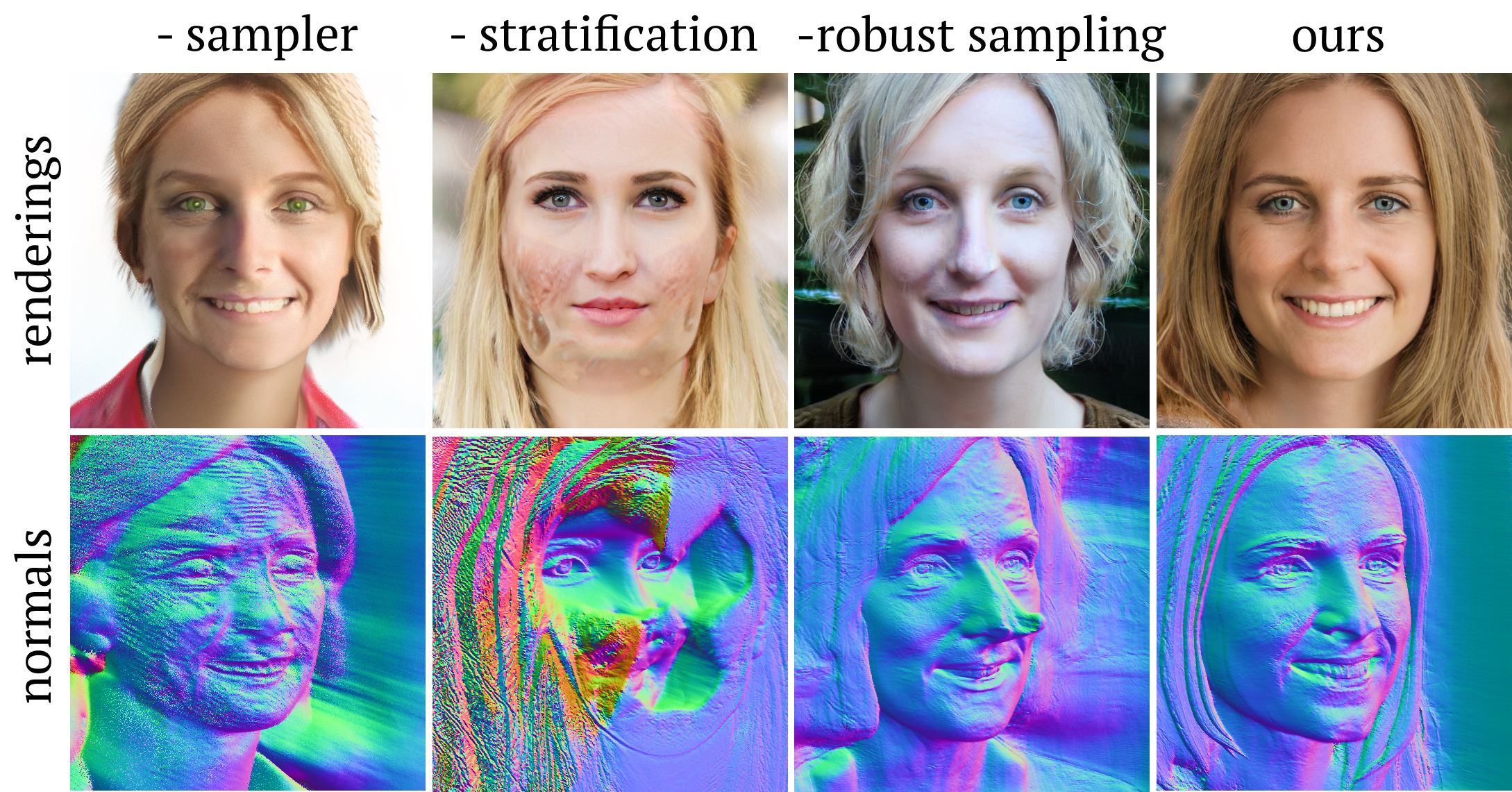}
    \caption{Qualitative comparisons for ablation study.}
    \label{fig:ablation}
    \vspace{-0.3cm}
\end{figure}

Without our surface tightness regularization (Eq.~\ref{equation:beta}), the SDF surface may get fuzzy, resulting in a less clean surface (see Fig.~\ref{fig:beta}) and worse geometry scores (see Tab.~\ref{tab:ablation}). Without our sampler or stratification during sampling, the model cannot learn meaningful 3D geometry with limited depth budgets, creating degenerated 3D geometry as can be seen in Fig.~\ref{fig:ablation} and significantly worse FID-N. Without our robust sampling strategy, the sampling becomes more susceptible to slight errors in the sampler due to the high-frequency nature of the PDF (see Fig.~\ref{fig:pdf}), resulting in a noticeable drop in FID and occasionally creating floater geometry artifacts (see Fig.~\ref{fig:ablation}), while geometry scores remain similar.

\begin{table}[t!]
\centering
\setlength\tabcolsep{3pt} %
\begin{tabular}{c c c c c c}
\toprule
 & & Method & FID $\downarrow$ & FID-N$\downarrow$ & NFS$\uparrow$ \\
\midrule
\multirow{6}{*}{\rotatebox[origin=c]{90}{FFHQ-512}} & \multirow{4}{*}{\rotatebox[origin=c]{90}{w/o SR}} & Epigraf~\cite{eg3d2022} & 9.92$^{\dagger}$ & 67.33 & \textbf{33.95} \\
 & & Mimic3D~\cite{mimic} & \underline{5.37} & \underline{64.97} & 16.76 \\
 & & GRAM-HD~\cite{xiang2022gramhd} & 12.2$^{\dagger*}$ & - & - \\
 & & Ours & \textbf{4.97} & \textbf{60.76} & \underline{29.35} \\
\cmidrule{2-6}
 & \multirow{2}{*}{\rotatebox[origin=c]{90}{w SR}} & EG3D~\cite{eg3d2022} & \textbf{4.70} & \textbf{63.02} & 17.54 \\
 & & StyleSDF~\cite{orel2022stylesdf} & 11.19\textsuperscript{\dag} &  87.42 & \textbf{22.75} \\
 & & MVCGAN~\cite{zhang2022mvcgan} & 13.4$^{\dagger*}$ & - & - \\
\bottomrule
\end{tabular}
\caption{Quantitative comparison with baselines with and without super resolution (SR) on the FFHQ dataset. \textsuperscript{\dag} as reported in the previous works. \textsuperscript{*} indicates FID evaluated on 20k images.}
\label{table:FFHQ}
\vspace{-.3cm}
\end{table}

\begin{table}[t!]
\centering
\begin{tabular}{c c c c c}
\toprule
 & & Method & FID $\downarrow$  & NFS$\uparrow$ \\
\midrule
\multirow{6}{*}{\rotatebox[origin=c]{90}{AFHQ-512}} & \multirow{4}{*}{\rotatebox[origin=c]{90}{w/o SR}} & GRAM-HD~\cite{xiang2022gramhd} & 7.67$^{\dagger}$ & - \\
 & & GMPI~\cite{orel2022stylesdf} & 7.79$^{\dagger}$ & - \\
 & & Mimic3D~\cite{mimic} & 4.29 & 12.67 \\
 & & Ours & \textbf{4.23} & \textbf{21.89} \\

\cmidrule{2-5}
 & \multirow{2}{*}{\rotatebox[origin=c]{90}{w SR}} & EG3D~\cite{eg3d2022} & \textbf{2.77} & 14.14 \\
 & & StyleSDF~\cite{orel2022stylesdf} & 7.91 & \textbf{33.89} \\
\bottomrule
\end{tabular}
\caption{Quantitative results on AFHQv2 Cats. \textsuperscript{\dag} as reported in the previous works.}
\label{table:AFHQ}
\vspace{-.3cm}

\end{table}

\begin{table}[t]
\centering
\small
\begin{tabular}{l c c c}
\toprule
Method & FID (20)$\downarrow$ & FID (50)$\downarrow$ & FID (96)$\downarrow$ \\
\midrule
Mimic3D & 53.57 & 13.31 & 5.37 \\
EG3D & 193.75 & 36.82 & \textbf{4.70} \\
Ours & \textbf{5.28} & \textbf{4.97} & 4.97 \\
\bottomrule
\end{tabular}
\caption{FID comparison on FFHQ using various sample counts. The samples per pixel are given in the parentheses of the metric.}
\label{table:20spp}
\end{table}

\begin{table}
\centering
		\small
    \begin{tabular}{l c c c}
\toprule
      Method         & FID $\downarrow$ & FID-N $\downarrow$ & NFS $\uparrow$   \\ 
        \midrule
- Learned Sampler & 38.29 & 93.88 & \textbf{30.95} \\
- Stratification & 5.60 & 86.02 & 5.97 \\
- Robust Sampling & 5.67 & \underline{60.78} & 24.79 \\
- Beta Regularization & \underline{5.27} & 64.25 & 28.88 \\
Ours & \textbf{4.97} & \textbf{60.76} & \underline{29.35} \\
\bottomrule
\end{tabular}
\caption{Ablation study.}
\label{tab:ablation}
    \vspace{-0.3cm}
\end{table}

\vspace{-.1cm}
\section{Discussion}

\paragraph{Limitations and future work.}
While our method demonstrates significant improvements in 3D geometry generation, it may still exhibit artifacts such as dents in the presence of specularities, and cannot handle transparent objects such as lenses well. Future work may incorporate more advanced material formulations~\cite{boss2021nerd} and surface normal regularization~\cite{verbin2022refnerf}. 
While 3D GANs can learn 3D representations from single-view image collections such as FFHQ and AFHQ with casual camera labels~\cite{eg3d2022}, the frontal bias and inaccurate labels can result in geometry artifacts, especially on the side of the faces.  
Fruitful future directions may include training 3D GANs with large-scale Internet data as well as incorporating a more advanced form of regularization~\cite{poole2022dreamfusion} and auto-camera calibration~\cite{An_2023_CVPR_Panohead} to extend the generations to 360 degrees. Finally, our sampling-based acceleration method may be applied to other NeRFs.
\vspace{-.3cm}
\paragraph{Ethical considerations.}
While existing methods have demonstrated effective capabilities in detecting unseen GANs~\cite{Corvi_diffusion_detection_icassp2023}, our contribution may remove certain characteristics from generated images, potentially making the task of detection more challenging. Viable solutions include the authentication of synthetic media~\cite{cai,c2pa,prashnani2023avatar}.   
\vspace{-.3cm}
\paragraph{Conclusion.}
We proposed a sampler-based method to accelerate 3D GANs to resolve 3D representations at the native resolution of 2D data, creating strictly multi-view-consistent images as well as highly detailed 3D geometry learned from a collection of in-the-wild 2D images. We believe our work opens up new possibilities for generating high-quality 3D models and synthetic data that capture in-the-wild variations and for enabling new applications such as conditional view synthesis.

\section*{Acknowledgements}
We thank David Luebke, Tero Karras, Michael Stengel, Amrita Mazumdar, Yash Belhe, and Nithin Raghavan for feedback on drafts and early discussions. 
Koki Nagano was partially supported by DARPA’s Semantic Forensics (SemaFor) contract (HR0011-20-3-0005). The views and conclusions contained in this document are those of the authors and should not be interpreted as representing the official policies, either expressed or implied, of the U.S. Government. This work was funded in part by an NSF Graduate Fellowship, ONR Grant
N00014-23-1-2526, and the Ronald L. Graham Chair. Manmohan Chandraker acknowledges support of of NSF IIS 2110409. Distribution Statement ``A'' (Approved for Public Release, Distribution Unlimited).

{
    \small
    \bibliographystyle{ieeenat_fullname}
    \bibliography{main}
}

\end{document}


\clearpage
\setcounter{page}{1}
\maketitlesupplementary

In this supplement, we first provide additional visual results (Sec.~\ref{sec:additional_results}) and additional evaluations (Sec.~\ref{sec:evaluation}). We follow with details of our implementation (Sec.~\ref{sec:implementation}) including the details of our adaptive sampling approach (Subsec.~\ref{subsec:trainsample}). We discuss experiment details (Sec.~\ref{sec:exp_details}) such as the details of our Normal-FID evaluation metric and baselines. We finally provide discussion (Sec.~\ref{sec:discussion}) including limitations of our work that may be addressed in future work. 
Please refer to the accompanying video, which contains additional visual results and comparisons. 

\section{Additional Qualitative Results}
\label{sec:additional_results}
\begin{figure*}[t!]
    \centering
    \includegraphics[width=0.95\linewidth]{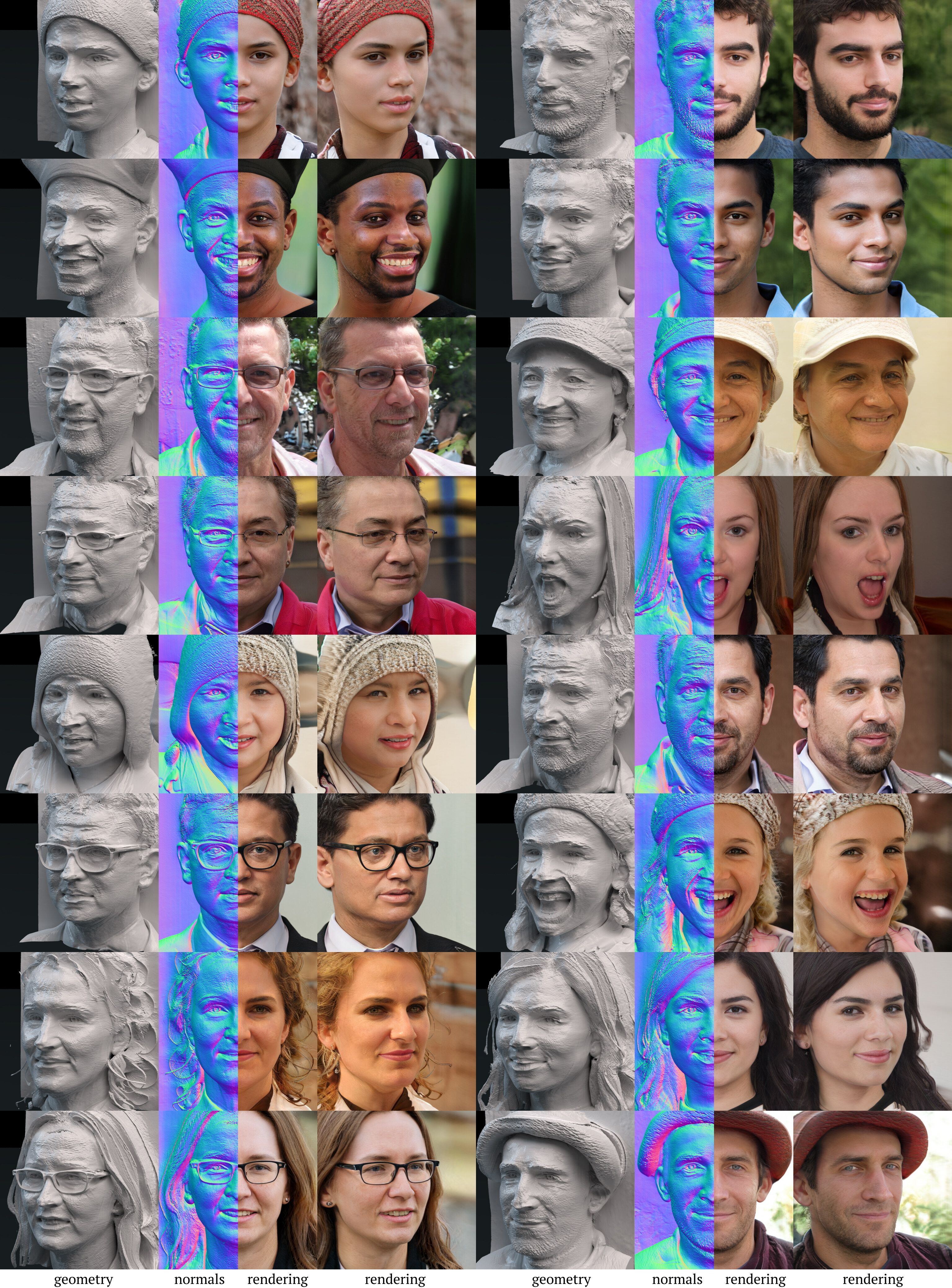}
    \caption{Curated samples from our FFHQ model.}
    \label{fig:curated}
\end{figure*}

\begin{figure*}[t!]
    \centering
    \includegraphics[width=0.95\linewidth]{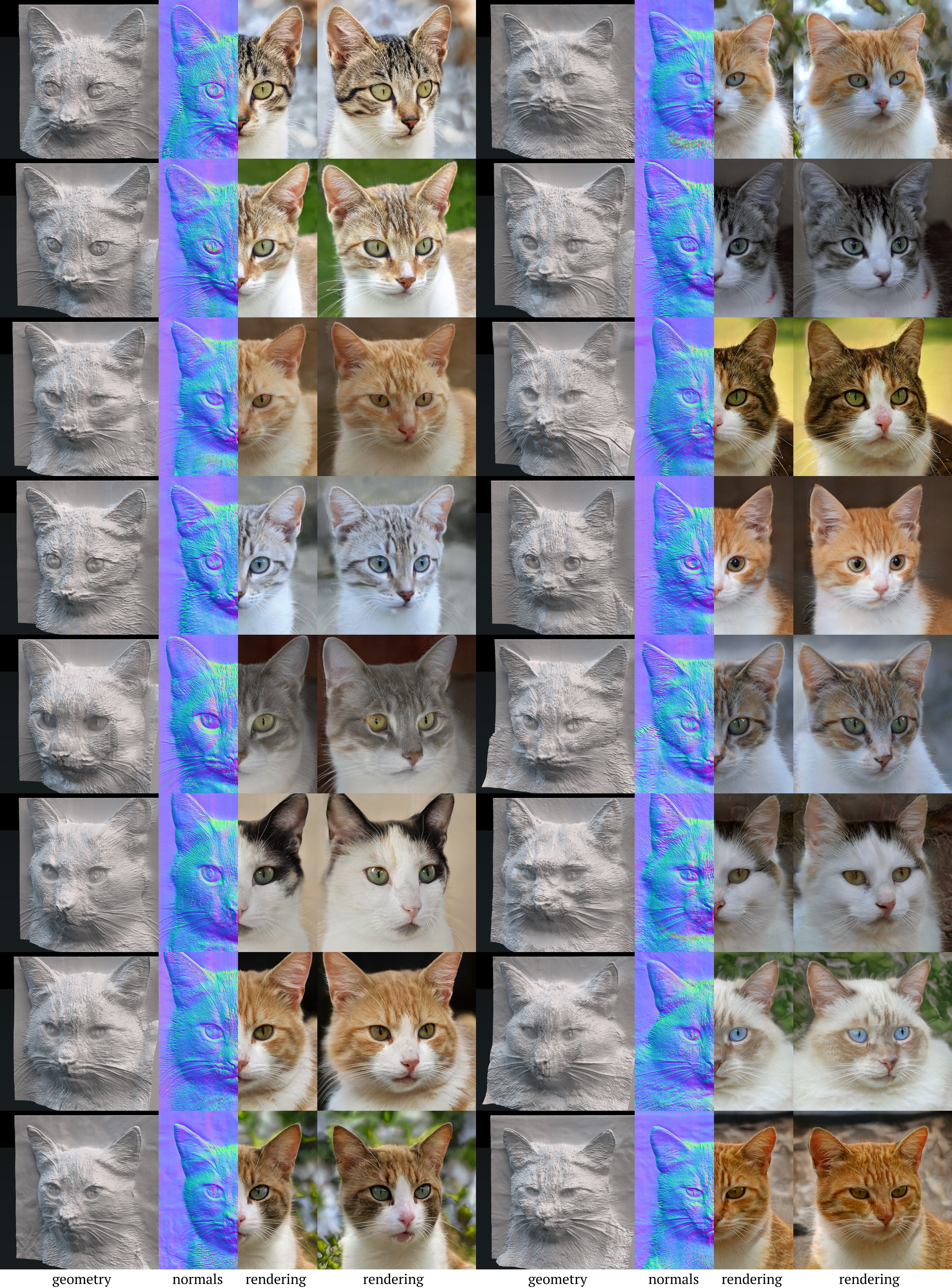}
    \caption{Curated samples from our AFHQ model.}
    \label{fig:curated_afhq}
\end{figure*}

\begin{figure*}[t!]
    \centering
    \includegraphics[width=0.95\linewidth]{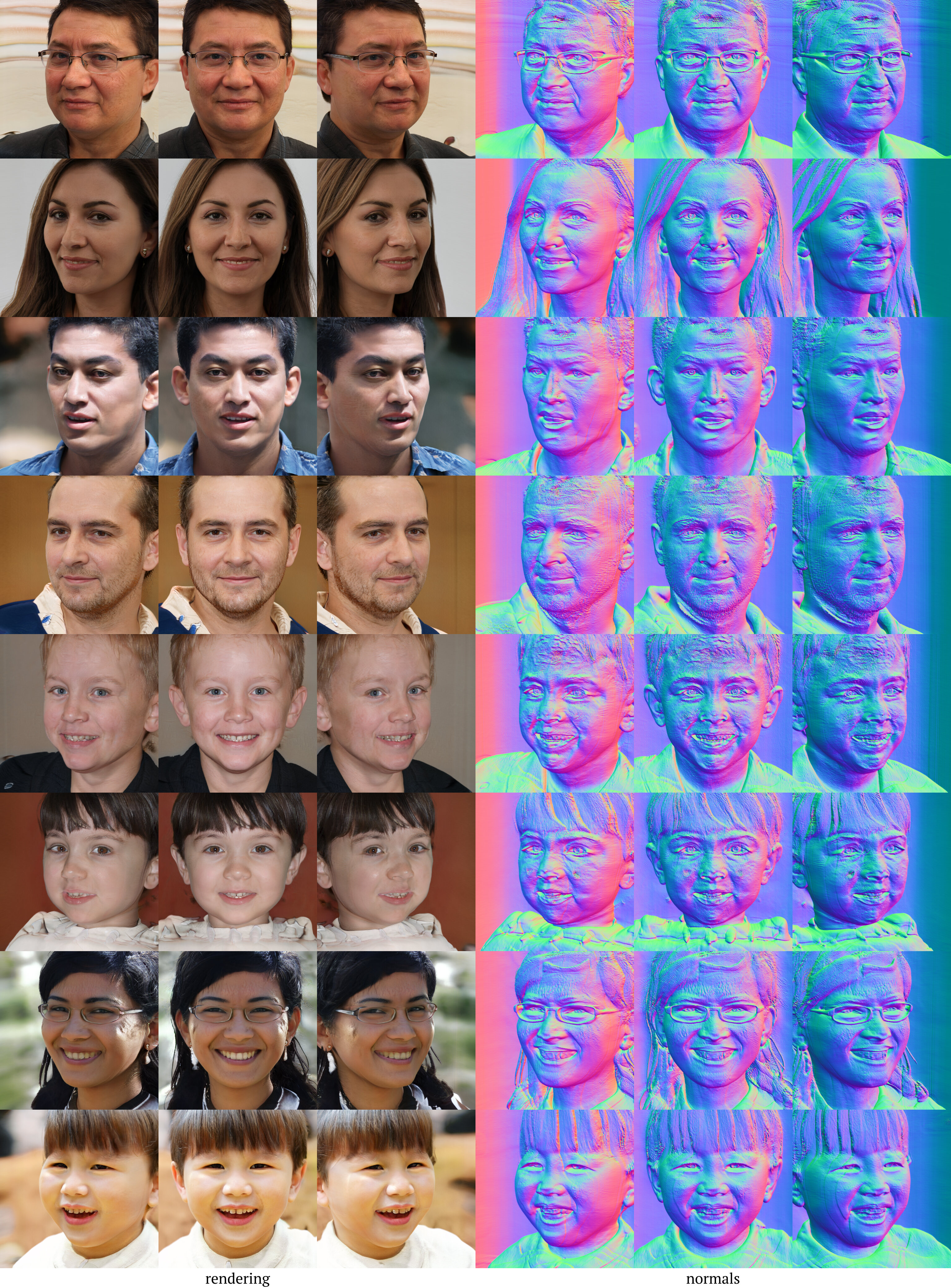}
    \caption{Uncurated (seeds 1-8) samples from our FFHQ model.}
    \label{fig:uncurated}
\end{figure*}

\begin{figure*}[t!]
    \centering
    \includegraphics[width=0.95\linewidth]{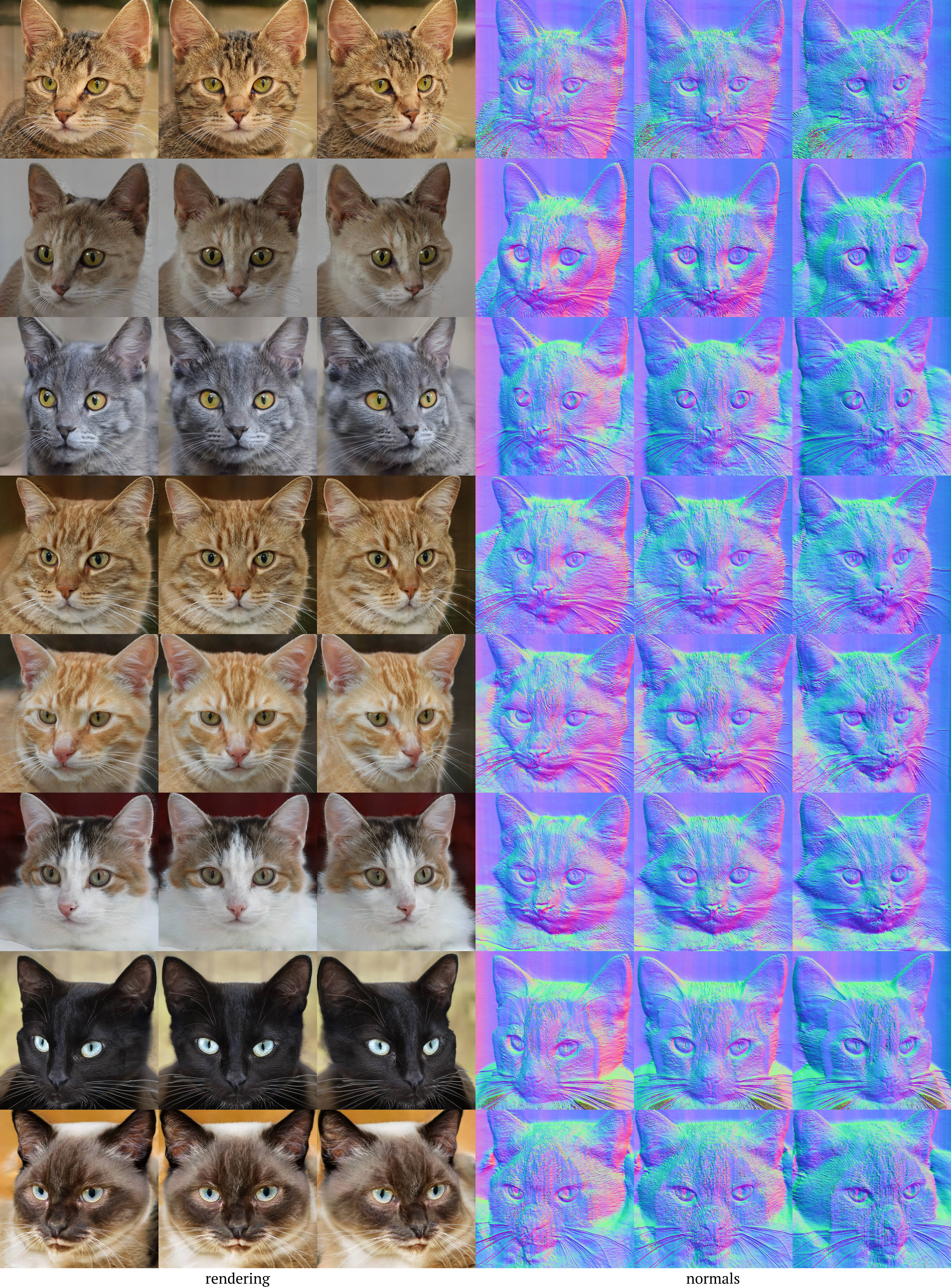}
    \caption{Uncurated (seeds 1-8) samples from our AFHQ model.}
    \label{fig:uncurated_afhq}
\end{figure*}

We first show both curated and uncurated results for both our FFHQ and AFHQ models. Curated FFHQ results can be seen in Fig.~\ref{fig:curated}. Please note the highly-detailed and variable facial expressions along with well-defined 3D accessories like hats and glasses. Long hair is not pasted onto the foreground, but rather retains a 3D aspect. Curated AFHQ results can be seen in Fig.~\ref{fig:curated_afhq}. Note the detailed textures of the cat geometry and the well-defined noses and ears. For unbiased presentation, we also show uncurated results (the first $8$ seeds) for FFHQ (Fig.~\ref{fig:uncurated}) and AFHQ (Fig.~\ref{fig:uncurated_afhq}). All results shown are with Truncation $=0.7$.

\section{Additional Evaluations}
\label{sec:evaluation}
\subsection{Comparing Sampling Methods at Various Sampling Counts}
\label{subsec:samplecount}
\begin{figure*}[t!]
    \centering
    \includegraphics[height=0.95\textheight]{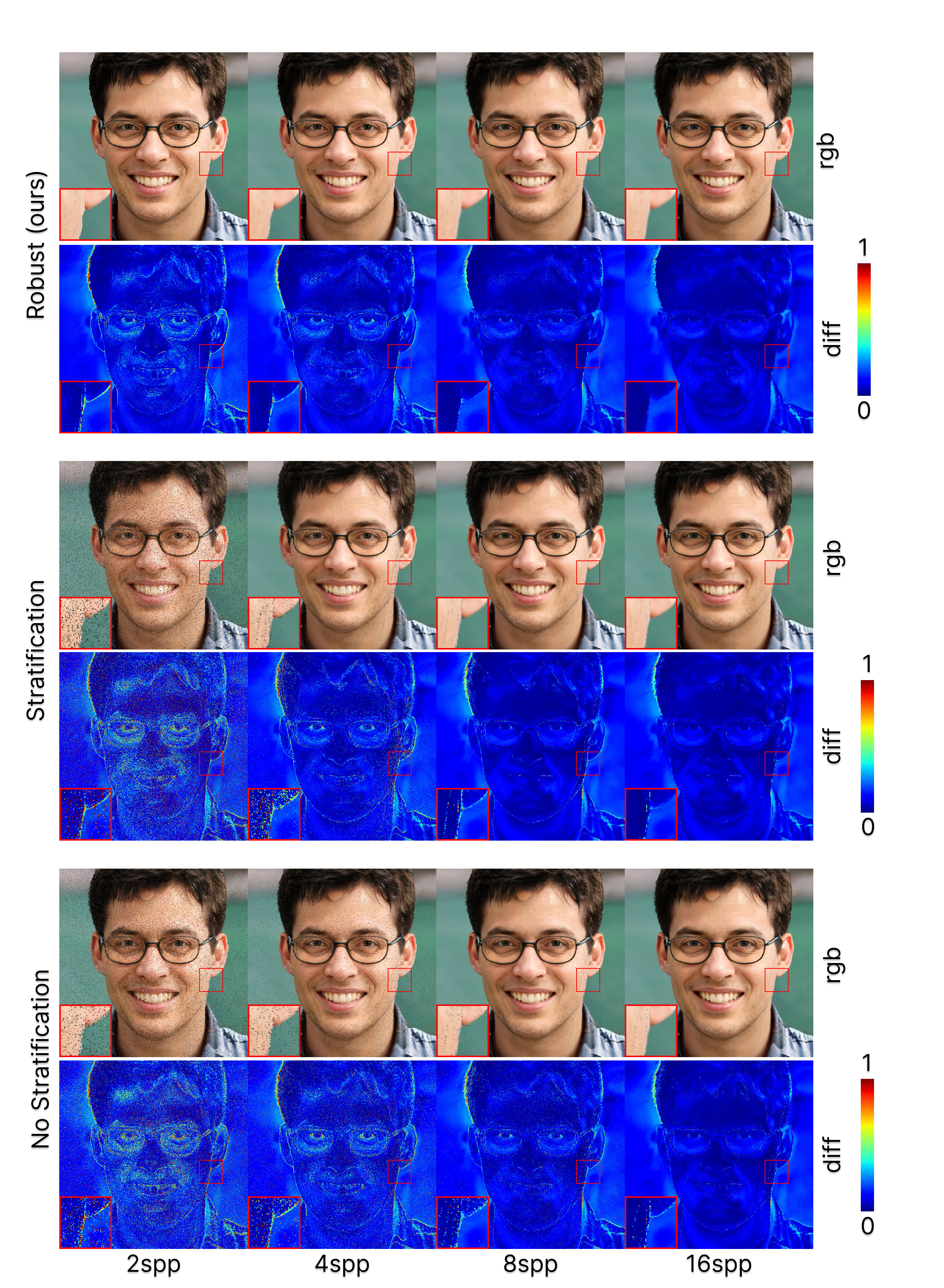}
    \caption{We show the rendering results of various sampling methods at different sample counts. We visualize both the rendering and a color map for the L2 error.}
    \label{fig:spp}
\end{figure*}

In this section, we show the robustness of our proposed sampling strategy from the predicted $\hat P_{512}$ at very low sample counts, in comparison to unstratified and stratified sampling methods. In Fig.~\ref{fig:spp}, we show our proposed \emph{robust} sampling method in comparison to unstratified and stratified inverse transform sampling. At very low samples per pixel (spp), our method vastly out performs the standard sampling technique. Please see the insets where our method can handle depth discontinuities without jagged artifacts even at $8$ samples per pixel. 

\begin{figure*}[t!]
    \centering
    \includegraphics[width=\linewidth]{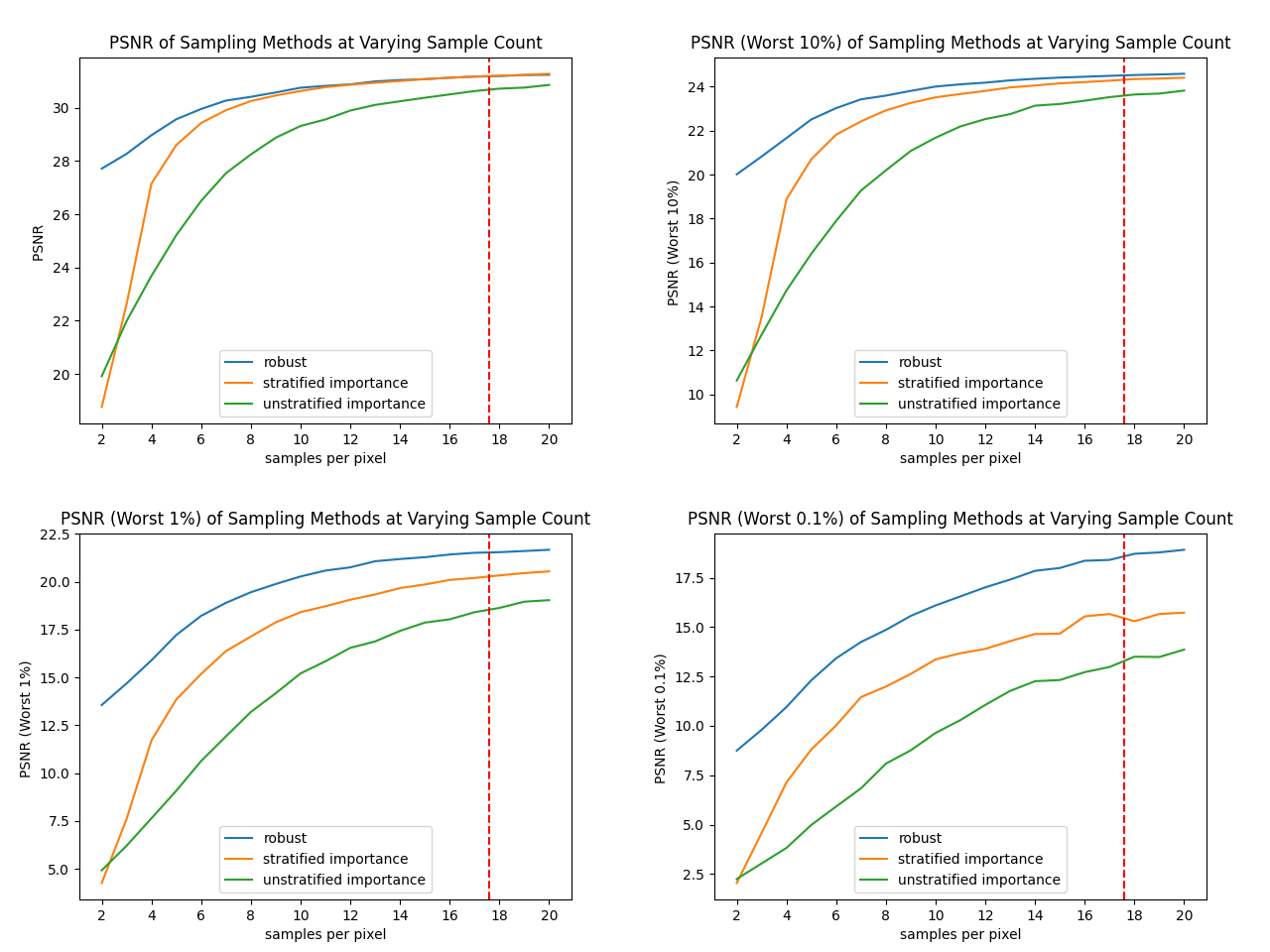}
    \caption{We show the PSNR for the worst percentage of pixels for all three sampling methods (subject is the same as Fig.\ref{fig:spp}). As seen in the charts, our method significantly outperforms the previous sampling methods at very low samples per pixel, e.g., $2$ samples. At higher sample counts, our proposed sampling method has a significantly better \emph{worst-case} result, e.g., for the worst $1\%$ or $0.1\%$ of pixels, as seen in the lower half. The importance of this is detailed in \ref{subsec:samplecount}. The red dotted line indicates the maximal number of samples during training that fit on one $80$gb A100 when rendering two images (per GPU).}
    \label{fig:spp_chart}
\end{figure*}

We also render a pseudo ground truth image using 384 (192 coarse and 192 importance) samples and compare the PSNR of various sampling methods in Fig.~\ref{fig:spp_chart}. Most importantly for GAN training, our method's \emph{worst-case} is significantly better than previous method's \emph{worst-case}. This is integral to GAN training, where the discriminator will always focus on the easiest attribute to discriminate. As sampling artifacts cannot be amended by $G$, the worst-case results dictate how well the GAN converges.

\subsection{Effectiveness of Adaptive Sampling}
\label{subsec:adaptive_sampling}
\begin{figure*}[t!]
    \centering
    \includegraphics[width=\linewidth]{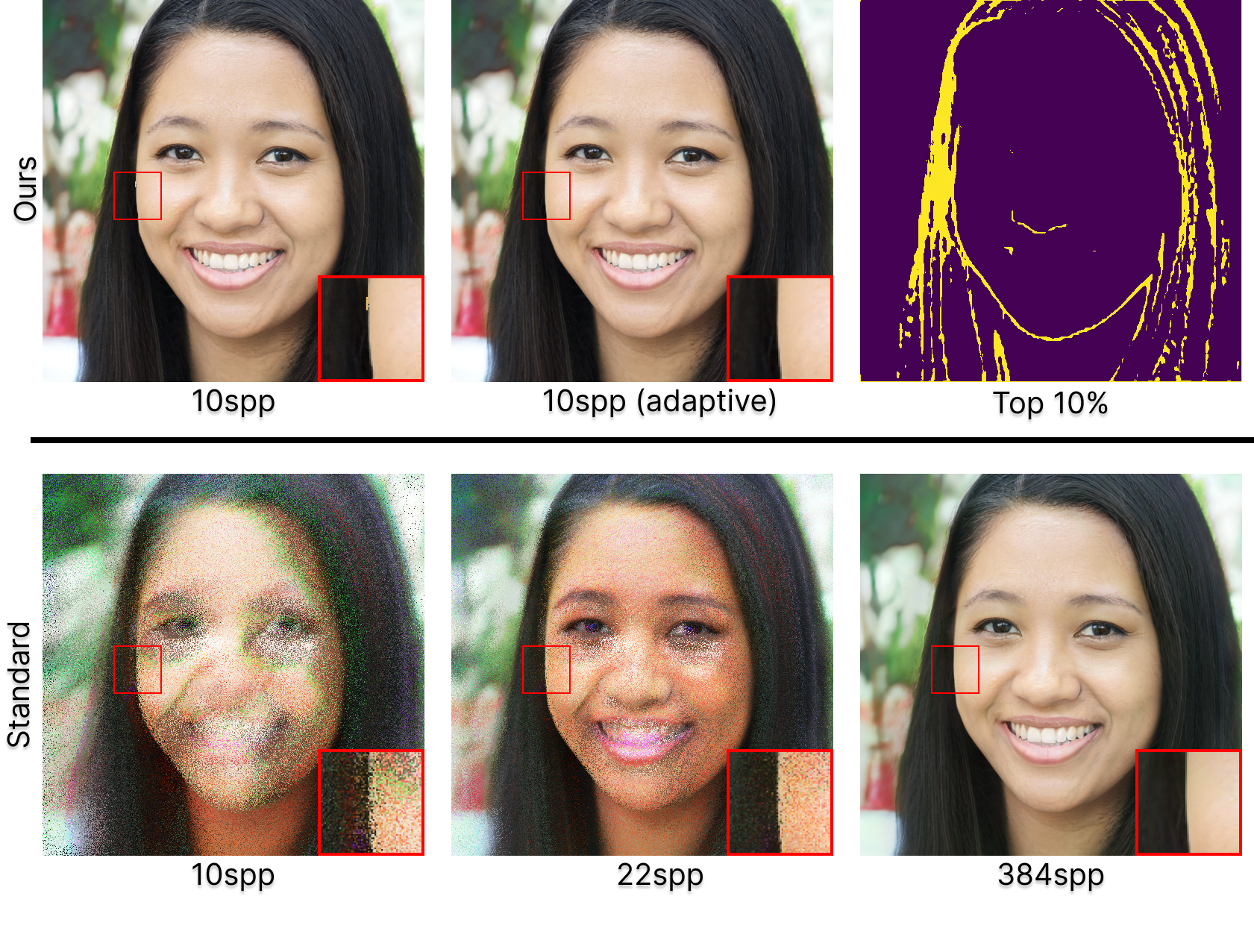}
    \caption{We visualize the effectiveness of our proposed adaptive sampling approach. In the top-left, we see that using 10 samples for all pixels results in jagged artifacts. Using the same number of samples, we allocate $9$ samples to $90\%$ of pixels and $19$ to the remaining $10\%$, which prevents these jagged artifacts. The top $10\%$ of pixels by the quantity computed in Subsec.~\ref{subsec:trainsample} is visualized in the top right. In comparison, we also show the standard method without the learned sampler with $10$ and $22$ samples in the bottom-left and bottom-middle, respectively. $22$ corresponds to $10$ samples along with the $12$ samples allocated for the initial probe $P_{128}$. Finally, we show the ground-truth rendering with $384$ samples (192 coarse and 192 importance) in the bottom right.}
    \label{fig:adaptive}
\end{figure*}

Fig.~\ref{fig:adaptive} demonstrates the effectiveness of our proposed adaptive sampling method compared to other baselines. By allocating a small portion of samples to uncertain regions (e.g., depth discontinuity; see the top right of Fig.~\ref{fig:adaptive}), our method can generate an artifact-free result even at the depth count budget of 10 samples per pixel (10spp) compared to the same spp without adative sampling (top left), which has jaggy artifacts around the depth discontinuity. Without our sampler (bottom row of Fig.~\ref{fig:adaptive}), the standard two-pass importance sampler~\cite{mildenhall2020nerf} results in significant artifacts. Please see Subsec.~\ref{subsec:trainsample} for the implementation details of our adaptive sampling.

\subsection{Single Image Reconstruction}
\label{subsec:inversion}
\begin{figure*}[t!]
    \centering
    \includegraphics[width=\textwidth,height=0.9\textheight,keepaspectratio]{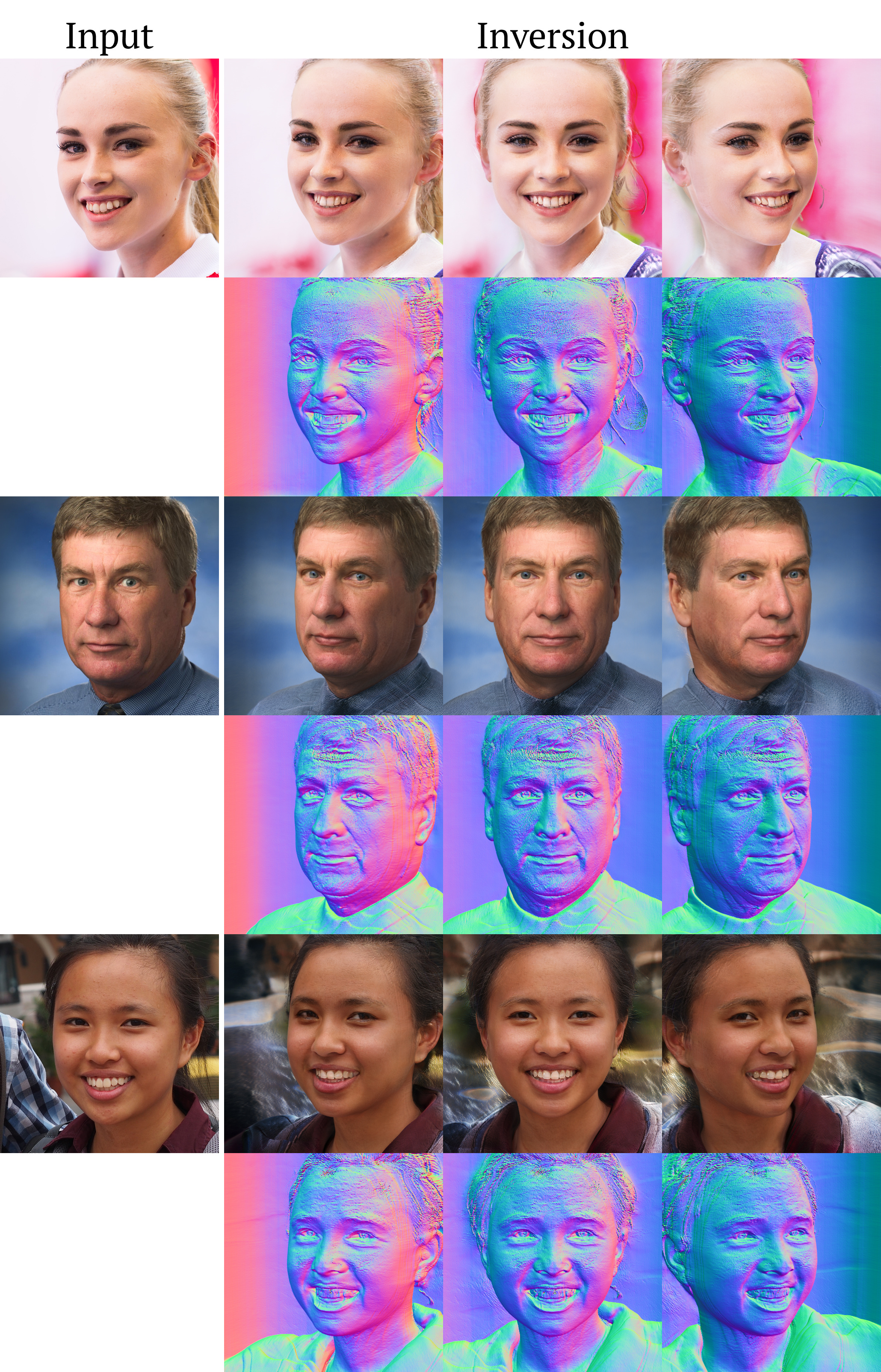}
    \caption{We showcase examples of single-view 3D reconstruction with our method. The left column shows the test image inputs, the right three columns show our inversion sample from three novel views.}
    \label{fig:inversion}
\end{figure*}

We additionally showcase an application of our method for single-view 3D reconstruction in Fig.~\ref{fig:inversion}. The learned prior enables high quality reconstruction of images and 3D geometry, despite the under constrained nature of the problem. We incorporate Pivotal Tuning Inversion (PTI) \cite{roich2021pivotal}, optimizing the latent code, camera, and noise buffers for 600 iterations, followed by optimization of the camera and generator weights for another 350 iterations with MSE and LPIPS~\cite{zhang2018perceptual} losses computed between the input view and rendering.

\section{Implementations Details}
\label{sec:implementation}
\subsection{Inference Details and Time}
\label{subsec:inference}
During inference, our method, by default, uses 17.6 depth samples (see Subsec.~\ref{subsec:trainsample} for details) at high resolutions in addition to samples from the low-resolution probe, which is equivalent to 12 samples at high resolutions; this results in 29.6 samples per ray at high resolutions. While our model learns view-dependent effects during training (see \textcolor{red}{Eq.~5} in the main PDF), we use a constant frontal-viewing condition during inference for our qualitative results. Rendering from cached triplanes runs at 4.5 FPS using plain PyTorch scripts on a single A100 GPU and requires $<$15GB of VRAM.

\subsection{Training Details}
\label{subsec:training}
In this section, we present the details of the training of our proposed model. For the schedule of the hyperparameters of the FFHQ model, please see Tab.~\ref{tab:hyperparameters}. We train the FFHQ model for $28.2$ million images with a batch size of $32$ images on $16$ $80$GB NVIDIA A100 GPUs, which takes about $11$ days. The AFHQ model is finetuned from this model with adaptive discriminator augmentation~\cite{Karras2020ada} for $1.2$ million images and R1 gamma value of $6$, and all other hyperparameters the same as in the end of the FFHQ training. 

\subsection{Network Details}
\label{subsec:implementation}
The architecture of the generator for $T'$ follows EG3D~\cite{eg3d2022} exactly, except doubling its capacity (channel base from $32768$ to $65536$). As mentioned in the main paper, we add three extra Synthesis Blocks from StyleGAN2~\cite{Karras2020stylegan2} applied to the channels of $T'$, in order to get the final triplane $T$--one for each of the orthogonal planes and each applied to one of the three slices of 32 channels of $T'$.

For the details of the architecture of the proposal network, please refer to Tab.~\ref{tab:samplermodel}. Slightly abusing notation, we have labelled the image of viewing directions corresponding to the target camera as $\phi_{128}$, parameterized as normalized vectors per pixel. The other inputs, $P_{128}$ (weights) and $I_{128}$ (image), follow the same notation as the main paper. 

For $\text{MLP}_\text{SDF}$, we embed the input 3D positions with the embedding from NeRF~\cite{mildenhall2020nerf} and $6$ frequencies (sampled in logspace). The architecture of this network is given in Table~\ref{tab:sdfnetwork}. The first two components of the $66$ dimensional output correspond to the SDF value $s$ and pre-activated $\beta_\text{pre}$. We map to the variance with the following equation: $\beta = 0.01 + \text{Tanh}(2\cdot\beta_\text{pre}) (0.01 - 0.0001)$. This activation ensures that the variances are approximately $0.01$ at the beginning of training and prevents them from becoming negative. 

We also show the details of $\text{MLP}_c$ in Table~\ref{tab:colornetwork}. The positional encoding of the viewing direction is $2$ frequencies sampled in logspace. We finally map the $3$ output components $c_\text{pre}$ to the RGB value as $\text{Sigmoid}\cdot(c_\text{pre})(1 + 0.002) - 0.001$.

\begin{table*}[h]
\renewcommand{\arraystretch}{1.2} %
\centering
\begin{tabular}{|c|c|c|c|c|}
\hline
\textbf{Layer} & \textbf{Type} & \textbf{Input} & \textbf{Activation} & \textbf{Dimension} \\
\hline
Input 0 & Input & XYZ positions & - & 3 \\
Input 1 & Input & Triplane features & - & 32 \\
\hline
1 & PosEnc & Input 0 & - & 39 \\
2 & Concatenation & Input 1, Layer 1 & - & 71 \\
3 & Linear & Concatenation & Softplus &  128 \\
4 & Linear & Layer 3 & Softplus & 128 \\
5 & Linear & Layer 4 & - & 66 \\
\hline
\end{tabular}
\caption{Architecture of the MLP$_\text{SDF}$ network with embedding.}
\label{tab:sdfnetwork}
\end{table*}

\begin{table*}[h]
\renewcommand{\arraystretch}{1.2} %
\centering
\begin{tabular}{|c|c|c|c|c|}
\hline
\textbf{Layer} & \textbf{Type} & \textbf{Input} & \textbf{Activation} & \textbf{Dimension} \\
\hline
Input 0 & Input & Viewing Directions $\phi$ & - & 3 \\
Input 1 & Input & $f_\text{geo}$ & - & 64 \\
\hline
1 & Embedding & Input 0 & - &  15 \\
2 & Concatenation & Input 1, Layer 1 & - & 79 \\
3 & Linear & Layer 2 & Softplus & 64 \\
4 & Linear & Layer 3 & - & 3 \\
\hline
\end{tabular}
\caption{Architecture of the MLP$_c$ network with embedding for viewing direction.}
\label{tab:colornetwork}
\end{table*}

\begin{table*}[htbp]
\begin{tabular}{@{}lll@{}}
\toprule
Hyperparameter                      & Number of Images (in millions) & Value \\ \midrule
\multirow{3}{*}{R1 Gamma}           & 0-18           & 1     \\
                                    & 18-25         & 4     \\
                                    & 25 onwards       & 2     \\ \midrule
\multirow{3}{*}{Neural Rendering Resolution} & 0-10           & 64    \\
                                    & 10-18         & 128 (linearly increased over 1m images)   \\
                                    & 18 onwards       & 512 (linearly increased over 0.2m images)   \\ \midrule
\multirow{2}{*}{$\beta_\text{target}$}               & 0-10           & 0.01  \\
                                    & 10 onwards     & 0.001 (linearly decreased over 1m images) \\ \midrule
\multirow{2}{*}{Learning Rate Multiplier for MLP$_\text{c}$} & 0-25          & 2     \\
                                    & 25 onwards     & 1     \\ \midrule
\multirow{2}{*}{Render with Predicted Distributions} & 0-17          & No    \\
                                    & 17 onwards     & Yes   \\ \midrule
\multirow{2}{*}{Supervise Predicted Distributions}   & 0-16          & No    \\
                                    & 16 onwards     & Yes   \\ \bottomrule
\end{tabular}
\centering
\caption{Schedule of hyperparameters given in millions of images the discriminator has seen during training. We train for 28.2m images total.}
\label{tab:hyperparameters}
\end{table*}

\begin{table*}[h]
\renewcommand{\arraystretch}{1.2} %
\centering
\begin{tabular}{|c|c|c|c|c|c|}
\hline
\textbf{Layer} & \textbf{Type} & \textbf{Activation} & \textbf{Upsample} & \textbf{Input Source(s)} & \textbf{Dimension} \\
\hline
Input 0 & Input & - & - & \( P_{128} \) & 191 x 128 x 128 \\
Input 1 & Input & - & - & \( I_{128} \) & 3 x 128 x 128 \\
Input 2 & Input & - & - & \( \phi_{128} \) & 3 x 128 x 128 \\
\hline
1 & Concatenation & - & - & Inputs 0-2 & 197 x 128 x 128 \\
2 & Conv2D & ReLU & No & Layer 1 & 256 x 128 x 128 \\
3 & Conv2D & ReLU & No & Layer 2 & 256 x 128 x 128 \\
4 & Conv2D & ReLU & No & Layer 3 & 256 x 128 x 128 \\
5 & Conv2D & ReLU & No & Layer 4 & 256 x 128 x 128 \\
6 & Conv2D & ReLU & Yes & Layer 5 & 256 x 256 x 256 \\
7 & Conv2D & ReLU & No & Layer 6 & 256 x 256 x 256 \\
8 & Conv2D & ReLU & No & Layer 7 & 256 x 256 x 256 \\
9 & Conv2D & ReLU & Yes & Layer 8 & 256 x 512 x 512 \\
10 & Conv2D & None & No & Layer 9 & 191 x 512 x 512 \\
11 & BilinearUpsample & - & Yes & Input 0 & 191 x 512 x 512 \\
12 & Conv2D & ReLU & No & Layer 10 + Layer 11 & 256 x 512 x 512 \\
13 & Conv2D & ReLU & No & Layer 12 & 256 x 512 x 512 \\
14 & Conv2D & ReLU & No & Layer 13 & 256 x 512 x 512 \\
15 & Conv2D & Softmax & No & Layer 14 & 191 x 512 x 512 \\
\hline
\end{tabular}
\caption{Architecture of our proposal network.}
\label{tab:samplermodel}
\end{table*}

\subsection{Details of Adaptive Sampling}
\label{subsec:trainsample}
As mentioned in the main paper, we use a proxy for the variance to adaptively allocate more samples to more difficult pixels. Specifically, considering the predicted high-resolution distributions $\hat P_{512}$, for each distribution, we compute a scalar value to dictate how many samples to allocate. We operate under the simplified assumption that we will allocate $16$ samples to $90\%$ of pixels, and $32$ to the remaining $10\%$, resulting in total 17.6 samples per ray. To compute \emph{which} pixels receive more samples, we compute a proxy for the variance. 

To do so, for a given predicted distribution $p$, we compute the leftover probability mass after removing the largest $16$ bins. Precisely, we consider the set of non-repeating nonnegative integers less than $192$,
\begin{equation*}
    S = \left\{ (z_1, \ldots, z_{16}) \mid z_i \in \mathbb{Z}_{\geq0}, z_i < 192, z_i \neq z_j~\forall~ i \neq j \right\}.
\end{equation*}
We then find 
\begin{equation*}
S_{\text{max}} = \underset{(z_1, \ldots, z_{16}) \in S}{\text{max}} \, \sum_{i=1}^{16} p_{z_i}.
\end{equation*}
The final scalar $1-S_{\text{max}}$ is the leftover probability mass after removing the largest $16$ bins. We choose the $10\%$ of pixels from $\hat P_{512}$ which have maximized this quantity. A visualization of these pixels is given in Fig.~\ref{fig:adaptive}. We can see that they are most concentrated on the depth discontinuities where the distributions may not be unimodal. Adaptively allocating samples in this manner allows us to accurately render the most challenging pixels without wasting too many samples on the ``easier'' distributions. As can be seen in Fig.~\ref{fig:adaptive}, using the same total sample count, we can avoid jagged and inaccurate renders. For illustration purposes, we first show an example where all pixels are rendered with $10$ samples (top-left of Fig.~\ref{fig:adaptive}) and then with $9$ samples for $90\%$ of pixels, and $19$ samples for the remaining $10\%$, resulting in an average sample count of $10$ (top-middle of Fig.~\ref{fig:adaptive}). The samples to which we allocate more samples are visualized (top-right of Fig.~\ref{fig:adaptive}). We compare to varying sample counts with standard sampling in the bottom row of Fig.~\ref{fig:adaptive}.

\subsection{Details of Stratified Sampling}
\label{subsec:stratified}
As discussed in subsection \textcolor{red}{4.4}, we compute the robust distribution $q$ from the predicted distribution $\hat p$ from the proposal network. Let $I=\{i\in\mathbb{Z}:q_i>0\}$ denote the set of non-zero bins each with equal probability (as in the main paper). For stratified sampling, we partition the unit interval into $c=|I|$ strata. We assume we are given a sample budget $s>1$. We allocate $\lfloor \frac s {c} \rfloor$ samples to each of the $c$ strata. We then allocate one extra sample to the $(s \bmod c)$ bins with maximal $\hat p_i$. Note that as we allocate more samples to a particular stratum, the distance $\delta_i$ between adjacent intrastratum  samples shrinks, thus introducing no additional bias. In practice, we also clip the $\delta_i$ to the bin width to prevent outsized contributions from the endpoints of nonzero regions. For $s<c$, this is biased; however, in practice, due to our tightening regularization and adaptive allocation of samples, we almost always have $s\geq c$. Additionally, at extremely low spp, our method outperforms unbiased methods (see left column of Fig.~\ref{fig:spp}).

\section{Experiment details}
\label{sec:exp_details}
\subsection{Geometry Visualization}
\label{subsec:geometry_viz}
For geometry visualization, we extract iso-surface geometry using March Cubes~\cite{lorensen1987marching}. We use the voxel resolution of $512^3$ for comparisons and $1024^3$ for our main results. For SDF-based methods (ours), geometry is extracted from an SDF field at the $0$th level set. For NeRF-based methods (EG3D, Mimic3D, and Epigraf), the surface is extracted from the density field using the level set provided by the official script from the authors. We render these extracted models using Blender for visualization. To visualize normal maps, we derive the normal by taking the gradient of the SDF field for SDF-based methods and density field for NeRF-based methods with respect to positions. 

\subsection{Normal-FID}
\label{subsec:nfid}
\begin{figure*}[t!]
    \centering
    \includegraphics[height=0.9\textheight]{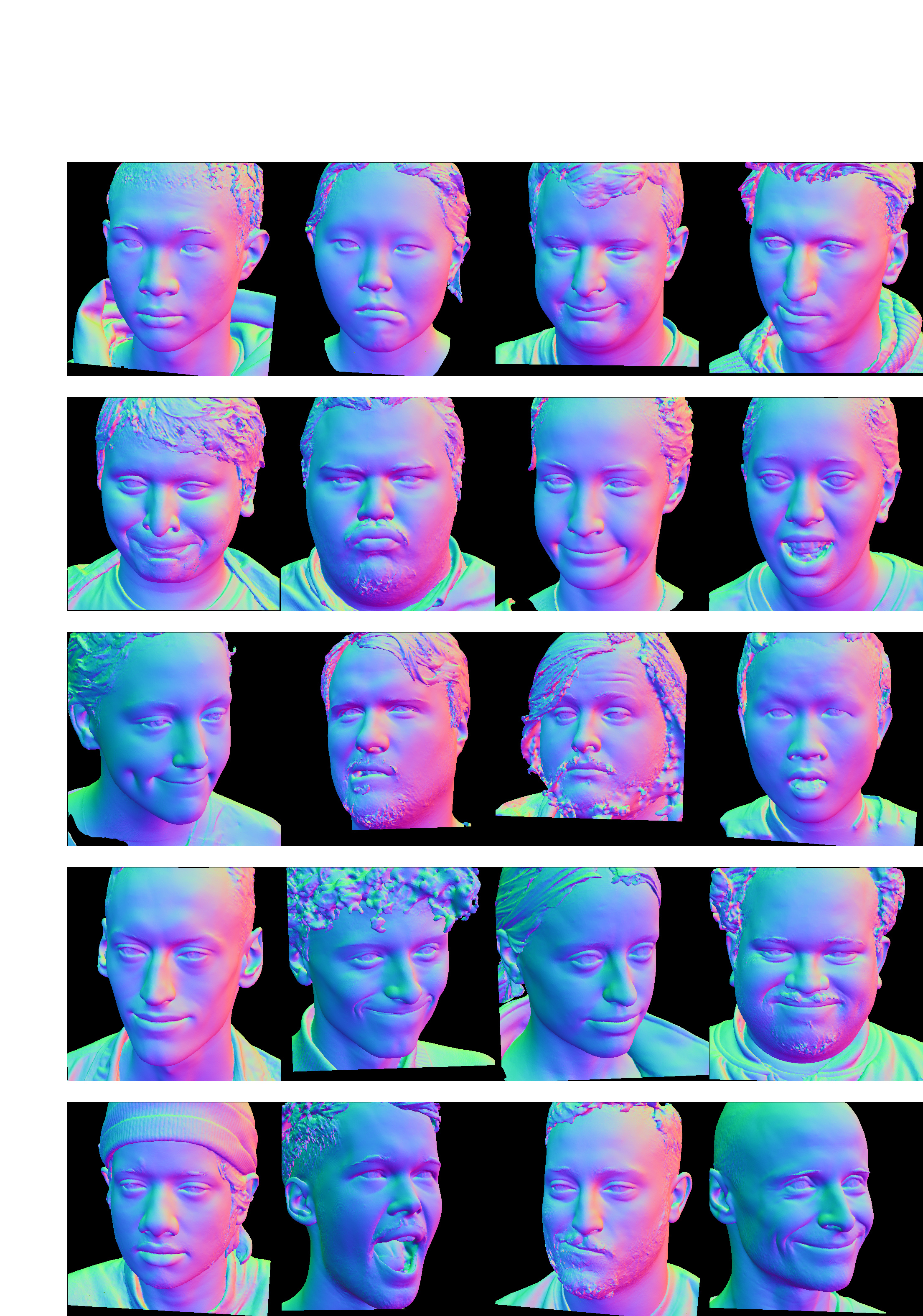}
    \caption{20 sample normal maps from~\cite{nphm} masked using Facer~\cite{facer}, from which we compute the N-FID score.}
    \label{fig:nphm}
\end{figure*}

As mentioned in the main text, we use normal maps extracted from the meshes of the NPHM~\cite{nphm} dataset. 255 subjects are scanned with highly variable expressions. We provide examples of these normal maps in Fig.~\ref{fig:nphm}. For Normal-FID computation, we ensure all coordinate conventions are consistent between baselines so that the color maps are likewise consistent. We sample all methods with truncation of $0.7$ (cutoff $=14$) due to the lack of diversity in the ground truth images (see Fig.~\ref{fig:nphm}). Using PyFacer~\cite{facer}, we also mask the background pixels to black. For EpiGRAF~\cite{epigraf}, we crop all sample images using HRN~\cite{hrn}.

\begin{figure*}[h!]
    \centering
    \includegraphics[width=0.95\linewidth]{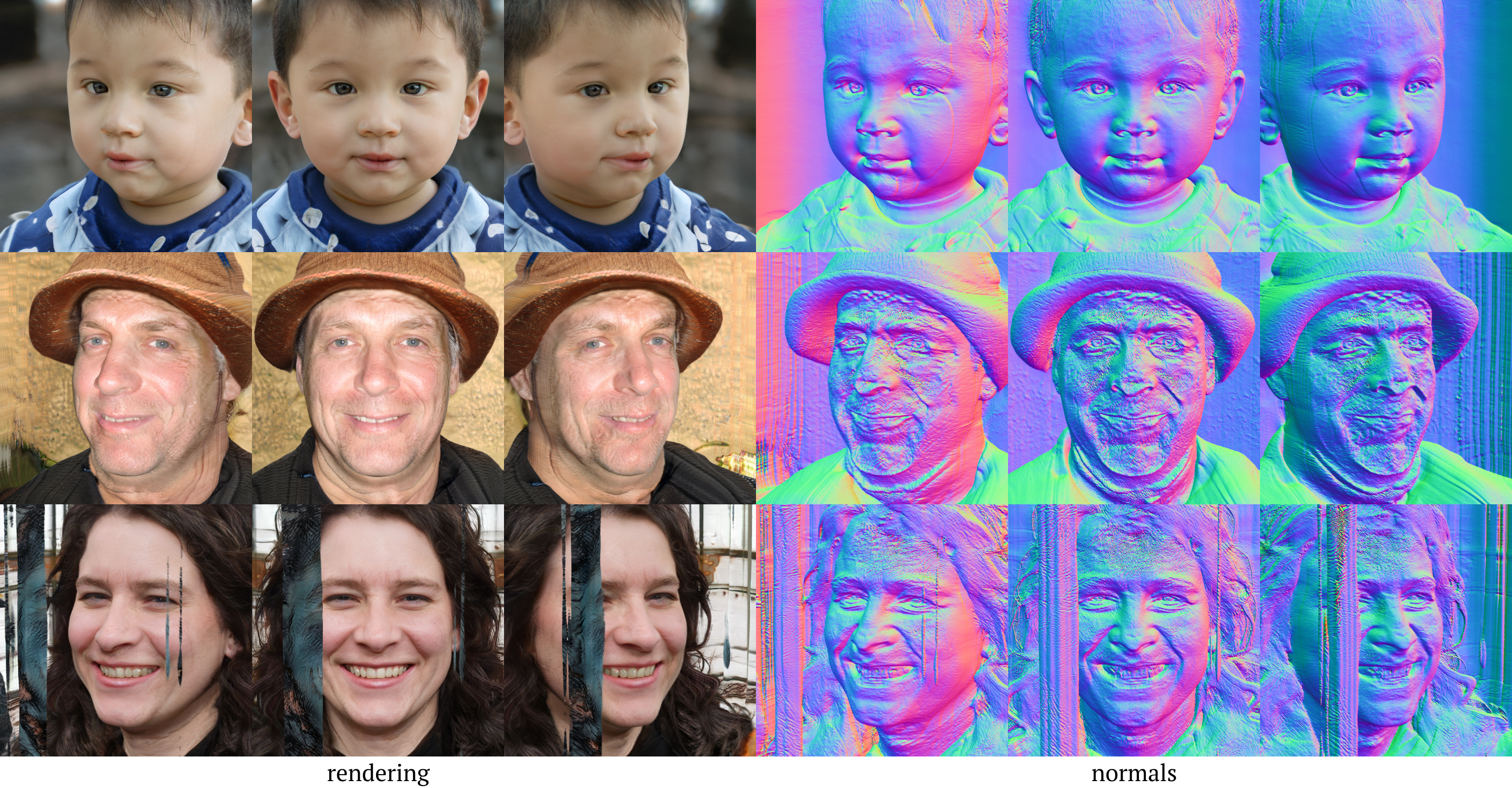}
    \caption{Three failure cases of our 3D generative model. First row shows seams on the side of the face; second row displays surface roughness to simulate specularity; and third row shows a rare phenomena where density appears close to the camera, occluding the subject.}
    \label{fig:limitation}
\end{figure*}

\subsection{Details of baseline methods}
\label{subsec:baseline_details}
For all the baselines, we use publicly released pre-trained models if they are available; otherwise, we quote the FID numbers from previous work. For Mimic3D~\cite{mimic}, Epigraf~\cite{epigraf} and EG3D~\cite{eg3d2022}, we used the corresponding publicly released models from the original authors for N-FID and non-flatness score computation; unless explicitly specified otherwise, we also use the provided default evaluation options for all methods. For EG3D, Mimic3d, Epigraf this is 48 samples for a coarse pass and 48 samples for a fine pass for two-pass importance sampling. For StyleSDF this is 24 samples per ray.

For StyleSDF~\cite{orel2022stylesdf}, we re-trained an FFHQ model at 512 resolutions and AFHQv2 cats-only model at 512 resolutions as they were not publicly available and used them for geometry evaluations. We train our StyleSDF geometry network on FFHQ using the publicly released code, for 200k iterations as recommended by the authors, on 8 A100 NVIDIA GPUs. For our StyleSDF geometry network on the AFHQv2 cats-only split, training with the provided AFHQ config from scratch was unstable and collapsed. Instead, we finetune the publicly released StyleSDF AFHQv2 all-animals geometry network on our cats-only split for 50k iterations and use that for evaluation.

\section{Discussion}
\label{sec:discussion}
\subsection{Limitation and Future Work}
\label{subsec:limitation}
We showcase three failure cases of our method in Fig.~\ref{fig:limitation}. In the first row, we see that there are seams in the side of the face in both the geometry and rendering. We hypothesize this issue may be related to the frontal camera bias in FFHQ and may be ameliorated by a more uniform sampling of cameras. In the second row, we see that in some samples with large amounts of specularity, the surface may become unnaturally rough, which may be remedied by additional regularization on the surface normal~\cite{verbin2022refnerf}. Finally, in the third row, we see a rare phenomena where density close to the camera occludes the subject. 

Future works may utilize more balanced datasets with larger coverage around the entirety of the face~\cite{An_2023_CVPR_Panohead}. Extending to the human body~\cite{ag3d} or more general classes~\cite{imagenet3d}, is also extremely interesting. Combining our approach with a 3D lifting approach~\cite{trevithick2023} using our method as 3D synthetic data, may allow high-fidelity geometry estimation from a single image.

{
    \small
    \bibliographystyle{ieeenat_fullname}
    \bibliography{main}
}